%% file: acl2025.tex
\pdfoutput=1
\documentclass[11pt]{article}

\usepackage[preprint]{acl}

\usepackage{times}
\usepackage{latexsym}
\usepackage[T1]{fontenc}

\usepackage[utf8]{inputenc}

\usepackage{microtype}
\usepackage{inconsolata}

\usepackage{graphicx}

\usepackage{booktabs}
\usepackage{hyperref}
\usepackage{multirow}
\usepackage{xcolor}

\include{macros}
\usepackage{subfigure}

\title{\PaperTitle}

\author{Weronika Łajewska\thanks{Work conducted during an internship at AWS AI Labs.}  \\
  \vspace{-0.3cm}University of Stavanger, \texttt{weronika.lajewska@uis.no}\\\AND
  Momchil Hardalov \quad Laura Aina \quad  Neha Anna John \quad Hang Su   \\
  Amazon \\
  \vspace{-0.3cm}\texttt{\{momchilh, eailaura, nehajohn, shawnsu\}@amazon.com} \\\AND
  \textbf{Llu\'{i}s M\`arquez\thanks{Work done while at Amazon}}  \\
  Technical University of Catalonia (UPC), \texttt{lluis.marquez@upc.edu}}

\begin{document}
\maketitle
\begin{abstract}
\input{sections/00-abstract}

\end{abstract}
\input{sections/01-intro}

\input{sections/02-related_work}

\input{sections/03-eval_framework}

\input{sections/04-initial_analysis}
\input{sections/05-results}
\input{sections/06-conclusions}

\bibliography{acl2025}

\clearpage
\appendix
\input{sections/07-appendix}

\end{document}

%% file: macros.tex
\usepackage{fontawesome}
\usepackage{xspace}
\usepackage[shortlabels]{enumitem}
\usepackage{makecell}
\usepackage{multirow}

\newcommand\PaperTitle{Understanding and Improving Information Preservation\\in Prompt Compression for LLMs}

\newcommand{\tocheck}[1]{\textcolor{black}{#1}}

%% file: sections/00-abstract.tex
Recent advancements in large language models (LLMs) have enabled their successful application to a broad range of tasks. However, in information-intensive tasks, the prompt length can grow fast, leading to increased computational requirements, performance degradation, and induced biases from irrelevant or redundant information. Recently, various prompt compression techniques have been introduced to optimize the trade-off between reducing input length and retaining performance. We propose a holistic evaluation framework that allows for in-depth analysis of prompt compression methods. We focus on three key aspects, besides compression ratio: (\emph{i})~downstream task performance, (\emph{ii})~grounding in the input context, and (\emph{iii})~information preservation. 
Using our framework, we analyze state-of-the-art soft and hard compression methods and show that some fail to preserve key details from the original prompt, limiting performance on complex tasks. By identifying these limitations, we are able to improve one soft prompting method by controlling compression granularity, achieving up to +23\% in downstream performance, +8 BERTScore points in grounding, and 2.7× more entities preserved in compression.
\tocheck{Ultimately, we find that the best effectiveness/compression rate trade-off is achieved with soft prompting combined with sequence-level training.}

%% file: sections/01-intro.tex
\section{Introduction}
\label{sec:intro}

Recent advancements in large language models have enabled their successful application to a broader range of tasks that require long-context input, detailed instructions, and in-context learning (ICL) demos~\citep{brown2020language,Wei:2022:NIPS, touvron2023llama,Yao:2024:NAACL, dubey2024llama,abdin2024phi}. The resulting prompts can become very long, especially for tasks requiring extensive contextual information. 
Even though state-of-the-art LLMs are able to process hundreds of thousands of tokens~\citep{anthopic2023claude21,munkhdalai2024leave,team2024gemini,yang2024qwen2,Intelligence2024}, providing them with long input context introduces computational inefficiencies~\citep{Shi:2023:ICML, Liu:2024:TACL}.

Recent research has focused on developing methods for prompt compression to reduce the length of the LLMs inputs while preserving the information needed to successfully complete the task~\cite{chang2024efficient,Li:2024:arXiv}. The two main approaches to compression are: \emph{soft-prompting}, which compresses the context into dense memory slots by learning continuous representations of the information in the latent space~\citep{Wingate:2022:EMNLP,Mu:2023:NIPS,Cheng:2024:NIPS}; and \emph{hard-prompting}, which operates on the surface form and involves removing unnecessary or low-information content, e.g.,~by pruning parts of the original text, or summarizing it~\citep{sennrich-etal-2016-neural,Jiang:2023:EMNLP,Pan:2024:ACL}. The resulting hard-compressed prompts still use natural language, but they may be less fluent. Soft compression methods allow for significantly higher compression rates~\citep{Li:2024:arXiva, Cheng:2024:NIPS}, as they operate in the embedding space. %
However, it is not easy to understand the information that is contained in soft prompts since, in contrast to hard prompting, these are not directly interpretable.

A limitation of the work done on prompt compression is that it focuses mostly on downstream performance on tasks like question answering (QA,~\citet{Mu:2023:NIPS}) or reading comprehension (RC,~\citet{Li:2024:arXiva}). We argue that downstream performance itself is insufficient to %
evaluate the quality and limitations of a 
compression method, and to understand its ability to preserve information from the original context.
To address this gap, we propose a holistic evaluation framework to analyze the performance of prompt compression approaches, which considers three key dimensions, beyond compression rate: (\emph{i})~downstream task performance, (\emph{ii})~grounding of the compressed context to the original text, and (\emph{iii})~type and amount of information preserved during compression. The framework includes the evaluation on various text generation tasks with different complexity and context lengths (from 850 to 6.5K tokens), including multi-hop reasoning, conversation QA, long document summarization, and mathematical reasoning.

We use our newly proposed framework to analyze a representative hard-prompting method, LLMLingua~\citep{Jiang:2023:EMNLP}, and two soft-prompting approaches, xRAG~\citep{Cheng:2024:NIPS} and PISCO~\citep{louis2025pisco}.
All these methods are limited in handling long-context scenarios, exhibiting performance drops across various generation tasks and an increased number of ungrounded responses. %
We also demonstrate that the target LLM struggles to accurately reconstruct the original context from soft prompts, particularly for xRAG, often hallucinating information and failing to regenerate crucial details.
Based on these observations, we 
explore modifications of xRAG, where we incorporate fine-grained data samples during 
pre-training and operate on more granular data 
at generation.
Our results show substantial gains: 23\% improvement on downstream tasks,
up to 8 BERTScore F1 points in response grounding, and 2.7× more entities preserved after compression.
Still, we see limitations in the ability to integrate information across separate units.
Our results show that this gap can be filled by soft prompting methods such as PISCO, which achieve the best trade-off between compression rate and effectiveness.

In summary, this paper provides the following contributions:
(\emph{i})~it reveals major limitations of state-of-the-art soft and hard prompt compression methods; (\emph{ii})~it advocates for a holistic framework to evaluate prompt compression methods beyond downstream task performance;
(\emph{iii})~it presents promising directions to address the limitations of soft prompting methods, showing that manipulating the granularity of the compressed content is key to improving performance.

%% file: sections/02-related_work.tex
\section{Related Work}
\label{sec:rel_work}

Prompt compression is a technique to improve the efficiency of LLMs by creating an approximate representation of a prompt with reduced size~\citep{wan2024efficient, Wingate:2022:EMNLP}. 
Approaches can be categorized into two main groups:
\textit{hard} prompting and \textit{soft} prompting~\citep{Li:2024:arXiv}.

\begin{table*}[t]
    \centering
    \resizebox{\linewidth}{!}{%
    \begin{tabular}{llllll}
        \toprule
        \# & \textbf{Dataset} & \textbf{Task} & \textbf{Input type} & \textbf{Output type} & \textbf{\# samples} \\
        \midrule
        1. &  HotpotQA~\citep{Yang:2018:EMNLP} & Multi-hop QA & Multiple documents & Short-form answer & 7,405 QA pairs (dev) \\
        2. &  QuAC~\citep{Choi:2018:EMNLP} & Conversational QA & Wikipedia section & Spans from the text & 863 QA pairs (validation) \\
        3. & TriviaQA~\citep{Joshi:2017:ACL} & RC with reasoning & Evidence documents & Short-form answer & 5,743 QA pairs (validation) \\
        4. & arXiv-summ.~\citep{Cohan:2018:NAACL} & Long-doc. summary & Scientific paper & Summary & 6,440 articles \\
        5. & GSM8K~\citep{Cobbe:2021:arXiv} & Math reasoning & Math problem & Answer with explanation & 1,000 questions (test) \\
        \bottomrule
    \end{tabular}
    }
    \caption{Overview of the datasets used for evaluating prompt compression methods. 
    }
    \label{tab:datasets}
\end{table*}

\paragraph{Hard Prompting} compression methods focus on producing a shorter text version of the input prompt, with minimal token usage to retain the key information. %
These methods operate at various granularity levels, from omitting low-information instruction content and ICL demonstrations to refining token selection~\citep{sennrich-etal-2016-neural, choi-etal-2024-reading}. Approaches include optimizing task definitions~\citep{yin-etal-2023-read}, dynamic compression allocation~\citep{Jiang:2023:EMNLP}, syntax-guided compression~\citep{yin-etal-2023-read} and demonstration selection via diversity-based sampling or relevance scoring~\citep{Yao:2024:NAACL, Gupta:2023:EMNLP}. 
Demonstration ordering can mitigate position bias~\citep{Jiang:2024:ACL}, while token classification ensures faithfulness to the original text~\citep{Pan:2024:ACL, Jiang:2023:EMNLP}.
However, hard prompting may disrupt grammar, introduce unfamiliar input distributions, and require re-encoding, impacting efficiency. Additionally, extreme compression is challenging, and compressing long inputs can be computationally expensive~\citep{Li:2024:arXiv}.

\paragraph{Soft Prompting} methods compress a given text into continuous representations~\citep{Lester:2021:EMNLP}. 
Soft prompts can achieve higher compression rates (i.e., fewer input tokens for the LLMs), but fall short in explainability--the content encoded in the soft prompt is not directly interpretable by humans.
Approaches like AutoCompressors recursively generate summary vectors
~\citep{Chevalier:2023:EMNLP}, while Gisting condenses prompts into transformer activations using virtual gist tokens~\citep{Mu:2023:NIPS}. ICAE~\citep{Ge:2024:ICLR} and 500xCompressor~\citep{Li:2024:arXiva} encode long contexts into compact memory slots: ICAE leverages a LoRA-adapted encoder and 500xCompressor uses key-value representations for richer information retention. xRAG~\citep{Cheng:2024:NIPS} projects dense retrieval embeddings into the LLM’s representation space via a trainable modality bridge. It employs a frozen embedding model as the encoder, with a lightweight adapter between the encoder and the decoder LLM as the only trainable component. Unlike other soft prompting methods, xRAG keeps both the encoder and decoder frozen, making it a modular and efficient solution for rapid development.
PISCO~\citep{louis2025pisco} is another compressor-decoder model relying on knowledge distillation from document-based questions. While xRAG minimizes the Kullback-Leibler divergence between the logits of the encoder and target LLM, PISCO is trained using sequence-level knowledge distillation from a teacher generator.

Most prompt compression methods are evaluated on downstream tasks, including classification~\citep{yin-etal-2023-read, Chevalier:2023:EMNLP}, natural language inference~\citep{Yao:2024:NAACL}, reasoning~\citep{chen-etal-2023-many}, QA~\citep{Jiang:2024:ACL, Cheng:2024:NIPS, Xu:2024:ICLR}, and summarization~\citep{Pan:2024:ACL, fei-etal-2024-extending}. 
However, downstream performance alone does not reveal a method's limitations or assess information loss, a key issue in prompt compression.
Reconstructing the original text from the compressed prompt has been proposed as a way to measure information preservation~\citep{Li:2023:arXiv, Wingate:2022:EMNLP, wang-etal-2024-context-former}, but standardized metrics for this evaluation are still lacking.
This work go beyond downstream performance by evaluating: (\emph{i})~the grounding of the generated responses in the original input, and (\emph{ii})~LLM's ability to reconstruct the input context from its compressed version. This approach enables us to study the quantity and nature of information preserved by compression methods.%

%% file: sections/03-eval_framework.tex
\section{Prompt Compression Evaluation Framework}
\label{sec:method}

\subsection{Benchmarking Data Design}
\label{subsec:benchmarking_data}

Currently, the main benchmarking of prompt compression methods~\citep{Jiang:2024:ACL} focuses on a limited range of tasks such as text classification  (e.g., MNLI~\citep{Yao:2024:NAACL}; SuperGLUE~\citep{Lester:2021:EMNLP,Chevalier:2023:EMNLP}), or extractive QA/RC (e.g.,~DROP,~\citealt{Gupta:2023:EMNLP}; RACE,~\citealt{Li:2024:arXiva}). 
Solving these tasks is often possible by only capturing a high-level topic representation or the main entities.
We argue that a more comprehensive evaluation is needed to truly cover the properties of the compression methods. Therefore, we adopt generation tasks containing much longer contexts, such as long-form QA based on multiple documents, long document summarization, and conversational search, with retrieval-augmented generation (RAG) based on multiple source documents. Table~\ref{tab:datasets} summarizes the target tasks characteristics. The list is not exhaustive and should evolve with advances in compression methods and LLMs.\footnote{ See Fig.~\ref{fig:datasets_context_length_distribution} in Appendix~\ref{app:datasets:eval} for context length distribution.} 

\begin{table*}[tp]
    \centering
    \resizebox{\textwidth}{!}{%
    \small
    \begin{tabular}{lcccccc}
        \toprule
        \textbf{Method} & \textbf{HotpotQA (EM)} & \textbf{HotpotQA* (EM)} & \textbf{arXiv-sum. (F1)} & \textbf{QuAC (F1)} & \textbf{TriviaQA (EM)} & \textbf{GSM8K (EM)} \\ %
        \midrule
        Mistral-7B & 0.664 & 0.772 & 0.834 & 0.869 & 0.773 & 0.477 \\ %
        Mistral-7B (no cont.) & 0.276 \textcolor{red}{(-58\%)} & 0.276 \textcolor{red}{(-64\%)} & --- & 0.834 \textcolor{red}{(-4\%)} & 0.590 \textcolor{red}{(-24\%)} & 0.440 \textcolor{red}{(-1\%)} \\ %
        xRAG & 0.297 \textcolor{red}{(-55\%)} & 0.374 \textcolor{red}{(-52\%)} & 0.803 \textcolor{red}{(-4\%)} & 0.838 \textcolor{red}{(-4\%)} & 0.691 \textcolor{red}{(-11\%)} & 0.336 \textcolor{red}{(-30\%)} \\ %
        PISCO & 0.318 \textcolor{red}{(-52\%)} & 0.589 \textcolor{red}{(-24\%)} & 0.818 \textcolor{red}{(-2\%)} & 0.861 \textcolor{red}{(-1\%)} & 0.738 \textcolor{red}{(-5\%)} & 0.393 \textcolor{red}{(-18\%)} \\ %
        LLMLingua & 0.306 \textcolor{red}{(-54\%)} & 0.696 \textcolor{red}{(-10\%)} & 0.805 \textcolor{red}{(-4\%)} & 0.846 \textcolor{red}{(-3\%)} & 0.727 \textcolor{red}{(-6\%)} & 0.305 \textcolor{red}{(-36\%)} \\ %
        \bottomrule
    \end{tabular}
    }
    \caption{Model performance on long-context datasets. Percentages in brackets show the relative \textcolor{red}{drop} from the Mistral-7B baseline, calculated as \emph{(score - mistral\_score) / mistral\_score}. ``No context'' (\emph{no cont.}) for GSM8K means removing ICL demos, while for arXiv-sum, the context is the document itself, making this setup not applicable.}
    \label{tab:res_baselines_downstream_tasks}
\end{table*}

\subsection{Compression Quality Evaluation}

We outline the following key dimensions for measuring the quality of the compression methods: (\emph{i})~downstream task performance, (\emph{ii})~response grounding, and (\emph{iii})~information preservation. 

\paragraph{Downstream Task Performance}
Following previous work, we use BERTScore (F1, \citet{Zhang:2020:ICLR}) for summarization and long-form QA, and Exact Match (EM) for reasoning and short-form QA (more details can be found in Appendix~\ref{app:experimental_setup:metrics}).

\paragraph{Grounding}
Grounding is another important dimension of compression quality, providing insights into a method's ability to preserve key information from the context to be used in the generated responses~\citep{Kim:2024:CML}.
We explored different approaches to automatically evaluate grounding, to find a generalizable metric across different long-context tasks.
We selected FABLES~\citep{Kim:2024:CML} as it produced results better aligned with human evaluation (see the discussion in Appendix~\ref{app:experimental_setup:grounding}).
FABLES first extracts a set of de-contextualized claims from the generated response and then rates the faithfulness of each claim given the evidence. Both steps are performed by prompting an LLM (\texttt{Claude 3 Haiku}). For each claim, we derive a score by comparing each 10-sentence context chunk against the claim and then taking the maximum score across chunks. The final score is the average across all claims in the response.

\paragraph{Information Preservation}

A key factor indicating the success of compression is the amount of main factual claims preserved from the original context.
To evaluate the capability of prompt compression methods to provide the target LLM with access to key information from the text, we look at the content from the original text that is preserved after compression. This evaluation is particularly interesting for soft prompting methods since compressed tokens are not interpretable by design. To capture the information preserved from the original text, we prompt the target LLM to reconstruct the content encoded in the soft prompt tokens. Then, the reconstructed text is compared with the original one using similarity metrics like ROUGE~\citep{Lin:2004:ACL} or BERTScore~\citep{Zhang:2020:ICLR}.

\paragraph{Compression Rates}

The last dimension that our framework monitors is the \emph{compression rate} -- the ratio between the size of the compressed and the original prompt. The computational cost of achieving compression is another important factor to be considered when comparing different methods.

%% file: sections/04-initial_analysis.tex
\section{Prompt Compression Methods Analysis}
\label{sec:analysis_prompt_compression}

For our experiments, we select two soft prompting and one hard prompting method, using the same target LLM -- \href{https://huggingface.co/mistralai/Mistral-7B-Instruct-v0.2}{\texttt{Mistral-7B Instruct v0.2}}~\citep{Jiang:2023:arXiv} -- across all setups, with the no-compression case as an upper bound.

\paragraph{Soft Prompting} 
We adopt xRAG~\citep{Cheng:2024:NIPS} and PISCO~\citep{louis2025pisco} as our soft prompting baselines. 
Both methods were shown to work in long context scenarios, making them the most promising starting point for our research. 
The input text in xRAG is encoded as a single token passed to the target LLM.
This representation is obtained from an encoder model, %
followed by a modality bridge  -- the only trainable component -- 
mapping the encoder's representation to the target LLM's embedding space. 
PISCO is a successor of xRAG containing two adapters around a backbone LLM: 1) an encoder adapter trained to compress input contexts into a set of embedding vectors, and 2) a decoder adapter providing an answer based on document embedding vectors and a query.

\paragraph{Hard Prompting}
We adopt \href{https://huggingface.co/spaces/microsoft/LLMLingua}{LLMLingua}~\citep{Jiang:2023:EMNLP} as our hard prompting baseline.
It dynamically allocates compression ratios to different prompt components (using a budget controller) while preserving semantic integrity. It then applies a token-level iterative pruning algorithm for fine-grained compression that accounts for conditional dependencies.\footnote{The compression budget of LLMLingua is 350 tokens.}
Following the LLMLingua setup~\citep{Jiang:2023:EMNLP}, the compressed prompt is provided as textual input to the target LLM.

\subsection{Downstream Task Performance}

We first evaluate the out-of-the-box compression methods on five diverse tasks (see Section~\ref{subsec:benchmarking_data}).\footnote{A reproducibility study confirmed that our results align with those reported in the original papers (see Appendix~\ref{app:reproducibility_study}).} Table~\ref{tab:res_baselines_downstream_tasks} compares the performance of Mistral-7B with/without input compression. %
We see that fully removing the context, such as background for QA or ICL examples for GSM8K, leads to large performance drops, confirming that the model's parametric memory alone is insufficient for these tasks. 

On this set of complex tasks, all compression methods result in sizeable performance losses (3\% to 55\% relative difference). The highest performance drop is observed on HotpotQA, as it requires
multi-hop reasoning and information aggregating. This indicates that applying compression removes crucial pieces of information needed to derive the correct answer. If we simplify the task, by retaining only the relevant paragraphs instead of the whole context (\emph{HotpotQA*}), we see an improvement of 8 points for xRAG and 39 for LLMLingua, 
indicating their difficulty with handling long, noisy contexts.

On the long document summarization tasks (\emph{arXiv-sum.}), differences between the base Mistral and the compression methods are smaller, with only a 4\% relative difference in F1 BERTScore. 
We attribute this 
to the nature of the task, i.e.,~summarizing very long texts (\textasciitilde5800 words) into concise abstracts (\textasciitilde200 words) requires focusing on high-level information rather than details.%

\emph{QuAC} and \emph{TriviaQA} tasks are less dependent on the details from the context. On these datasets, we see a smaller gap in performance
between the compressed and uncompressed inputs (5 EM points for LLMLingua and 8 points for xRAG). On the math reasoning task (\emph{GSM8K}), we find that both prompt compression methods ---in this case compressing ICL examples--- yield substantially worse performance than providing no demonstrations at all. This means that compressed ICL demonstrations are not well interpreted by the model, which makes the response generation even more difficult.

To assess the impact of the compression input granularity, we encode the input on context-, paragraph-, and sentence-level using xRAG tokens. We also run similar experiments with PISCO using separate sets of 8 embedding vectors per unit.
Results presented in Table~\ref{tab:res_baselines_xrag_per_tokens} indicate that compressing context into smaller segments does not improve performance for xRAG, likely because the tokens capture high-level topics without key details. However, PISCO trained with sequence-level knowledge distillation objective shows significant improvements on HotpotQA that uses longer contexts compared to the remaining two datasets when operating on smaller granularity. This suggests that enhancing the model to handle finer details is a worthwhile direction to explore for soft compression methods. %

\begin{table}[tp]
    \centering
    \resizebox{\columnwidth}{!}{%
    \setlength{\tabcolsep}{3pt}
    \begin{tabular}{llccc}
        \toprule
        \textbf{Method} & \textbf{Encodings} & \textbf{HotpotQA} & \textbf{HotpotQA*} & \textbf{TriviaQA} \\
        \midrule
        \multirow{3}{*}{xRAG} & 1 per context & 0.297 & 0.374 & 0.691\\
        & 1 per paragraph & 0.211 \textcolor{red}{(-30\%)} & 0.400 \textcolor{teal}{(+7\%)} & --- \\
        & 1 per sentence & 0.055 \textcolor{red}{(-81\%)} & 0.264 \textcolor{red}{(-29\%)} & 0.224 \textcolor{red}{(-68\%)} \\
        \midrule
        \multirow{3}{*}{PISCO} & 8 per context & 0.318 & 0.589 & 0.738 \\
        & 8 per paragraph & 0.512 \textcolor{teal}{(+61\%)} & 0.625 \textcolor{teal}{(+6\%)} & --- \\
        & 8 per sentence & 0.399 \textcolor{teal}{(+25\%)} & 0.559 \textcolor{red}{(-5\%)} & 0.719 \textcolor{red}{(-3\%)} \\
        \bottomrule
    \end{tabular}
    }
    \caption{Exact match scores for xRAG and PISCO responses with varying compression granularity.}
    \label{tab:res_baselines_xrag_per_tokens}
\end{table}

\subsection{Grounding}

We evaluate how prompt compression affects the faithfulness of LLM-generated text. 
Depending on the task, we have different grounding texts: source documents for summarization, and background sections for QA and RC.
We compare grounding scores for responses with and without compression to assess its impact.
Results in Table~\ref{tab:res_baselines_grounding} show that compression leads to a 30 points drop in groundedness score on HotpotQA, and around 50 points on QuAC and arXiv-summarization. 
This indicates that higher compression rates used in xRAG result in generating text that is less faithful to the context. The hallucinated content appears in the entire response/summary, including the first claim of the generated text, which intuitively contains the most important information (see Table~\ref{tab:res_baselines_grounding_avg_and_first} in Appendix~\ref{app:add_results:grounding}). As expected, responses generated with LLMLingua are less prone to hallucinations as the compressed input retains direct information from the original context. 
PISCO produces the most faithful responses to the source despite high compression rates, likely due to the use of sequence-level knowledge distillation from a teacher model.

\begin{table}[tp]
    \centering    
    \resizebox{\columnwidth}{!}{%
    \setlength{\tabcolsep}{3pt}
    \begin{tabular}{lcccccc}
        \toprule
        \textbf{Method} & \textbf{HotpotQA} & \textbf{HotpotQA*} & \textbf{arXiv-sum.} & \textbf{QuAC} & \textbf{TriviaQA} & \textbf{GSM8K} \\
        \midrule
        Mistral-7B & 0.80 & 0.75 & 0.97 & 0.93 & 0.78 & 0.50 \\
        xRAG & 0.52 & 0.57 & 0.39 & 0.45 & 0.73 & 0.42 \\
        PISCO & 0.59 & 0.76 & 0.74 & 0.63 & 0.84 & 0.48 \\
        LLMLingua & 0.45 & 0.75 & 0.62 & 0.49 & 0.72 & 0.44 \\
        \bottomrule
    \end{tabular}
    }
    \caption{FABLES grounding scores for responses generated with different methods,
    averaged over 5 random sets of 100 samples (stdev ($\sigma$) \(<0.04\)).}
    \label{tab:res_baselines_grounding}
\end{table}

\subsection{Information Preservation}
\label{subsec:results:infomration_preservation}
We reconstruct the original text from the compressed xRAG representation by prompting the target LLM to recreate the information encoded in the tokens. Using a subset of xRAG pre-training prompts (Table~\ref{tab:reconstruction_prompts} in Appendix~\ref{app:prompts}), we apply two scenarios: (\emph{i})~encoding the entire context 
in one xRAG token, and (\emph{ii})~encoding each context sentence into separate tokens. Similarly, for PISCO, we prompt the decoder to reconstruct the provided information, encoded on sample- or sentence-level, restoring as many details as possible. We hypothesize that compressing entire contexts into a single token risks significant information loss for longer, more complex inputs. Our evaluation of information preservation covers only soft prompting methods, not LLMLingua, as hard-prompting methods do not add new content and are fully interpretable.

\begin{table}[tp]
    \centering
    \resizebox{\columnwidth}{!}{%
    \setlength{\tabcolsep}{3pt}
    \begin{tabular}{lll|ccccc}
        \toprule
         &  &  & \multicolumn{4}{c}{\textbf{Sample Lengths}} & \textbf{Preserved} \\ 
        \textbf{Method} & \textbf{Data} & \textbf{Encodings} & \textbf{All} & \textbf{1-sent} & \textbf{5-sents} & \textbf{10-sents} & \textbf{Entities} \\
        \midrule
        \multirow{4}{*}{xRAG} & \multirow{2}{*}{Unseen}  & 1 per sample & 0.66 & 0.57 & 0.72 & 0.70 & 0.28 \\
        & & 1 per sent. & 0.42 & 0.57 & 0.38 & 0.31 & 0.19 \\
        \cmidrule{2-8}
        & \multirow{2}{*}{Seen} & 1 per sample & 0.65 & 0.50 & 0.73 & 0.73 & 0.25 \\ 
        & & 1 per sent. & 0.35 & 0.50 & 0.33 & 0.21 & 0.12 \\
         \midrule
        \multirow{2}{*}{PISCO} & 
         \multirow{2}{*}{Unseen} & 8 per sample & 0.89 & 0.91 & 0.89 & 0.87 & 0.49 \\
         & & 8 per sent. & 0.90 & 0.90 & 0.90 & 0.89 & 0.59 \\
        \bottomrule
    \end{tabular}
    }
    \caption{Information preservation results: BERTScore F1 between original and reconstructed context for different context lengths, and fraction of preserved entities.}
    \vspace{-0.5cm}
    \label{tab:res_xrag_reconstructing}
\end{table}

\begin{table*}[tp]
    \centering
    \resizebox{\textwidth}{!}{%
    \setlength{\tabcolsep}{3pt}
    \begin{tabular}{lcccccccccccccc}
        \toprule
        \multirow{2}{*}{\textbf{xRAG variant}} & \multicolumn{2}{c}{\textbf{HotpotQA (EM)}} & \multicolumn{2}{c}{\textbf{HotpotQA* (EM)}} & \multicolumn{2}{c}{\textbf{arXiv-sum. (F1)}} & \multicolumn{2}{c}{\textbf{TriviaQA (EM)}} & \multicolumn{2}{c}{\textbf{QuAC (F1)}} \\  %
         & \textbf{Cont.} & \textbf{Sent.} & \textbf{Cont.} & \textbf{Sent.} & \textbf{Cont.} & \textbf{Sent.} & \textbf{Cont.} & \textbf{Sent.} & \textbf{Cont.} & \textbf{Sent.} \\ %
        \midrule
        xRAG (Reproduced) & 0.286 & 0.014 & 0.375 & 0.174 & 0.775 & 0.760 & 0.696 & 0.115 & 0.829 & 0.843 \\ %
         w/ Sentence PT + FT & \textbf{0.313 \textcolor{teal}{(9\%)}} & 0.066 \textcolor{teal}{(371\%)} & 0.422 \textcolor{teal}{(13\%)} & 0.423 \textcolor{teal}{(143\%)} & \textbf{0.803 \textcolor{teal}{(3\%)}} & 0.739 \textcolor{red}{(3\%)} & \textbf{0.712 \textcolor{teal}{(2\%)}} & 0.361 \textcolor{teal}{(214\%)} & 0.833 \textcolor{teal}{(0\%)} & \textbf{0.855 \textcolor{teal}{(1\%)}} \\
         w/ Two-Step PT + FT & 0.277 \textcolor{red}{(-3\%)} & 0.145 \textcolor{teal}{(936\%)} & 0.406 \textcolor{teal}{(8\%)} & \textbf{0.462 \textcolor{teal}{(166\%)}} & 0.785 \textcolor{teal}{(1\%)} & 0.696 \textcolor{red}{(8\%)} & 0.685 \textcolor{red}{(-2\%)} & 0.521 \textcolor{teal}{(353\%)} & 0.834 \textcolor{teal}{(1\%)} & 0.823 \textcolor{red}{(-2\%)} \\ %
        \bottomrule
    \end{tabular}
    }
    \caption{Performance of the xRAG variants at different compression granularities, calculated on a subset of 1,000 examples, uniformly sampled from each dataset. Relative \textcolor{teal}{improvement}/\textcolor{red}{drop} is calculated wrt the reproduced xRAG.}
    \label{tab:res_xrag_variants_downstream_tasks}
\end{table*}

\begin{table}[tp]
    \centering
    \resizebox{\columnwidth}{!}{%
    \setlength{\tabcolsep}{3pt}
    \begin{tabular}{lll|cccccc}
        \toprule
        \textbf{Model} & \textbf{Data} & \textbf{Encodings} & \textbf{Avg.} & \textbf{PERSON} & \textbf{GPE} & \textbf{DATE} & \textbf{CARD.} & \textbf{ORG} \\ 
        \cmidrule{2-9}
        \multirow{4}{*}{xRAG} & \multirow{2}{*}{Unseen} & 1 per sample & 0.28 & 0.31 & 0.39 & 0.22 & 0.26 & 0.32 \\
        & & 1 per sentence & 0.19 & 0.21 & 0.25 & 0.14 & 0.13 & 0.24 \\
        \cmidrule{2-9}
        & \multirow{2}{*}{Seen}& 1 per sample & 0.25 & 0.16 & 0.38 & 0.24 & 0.38 & 0.21 \\
        & & 1 per sentence & 0.12 & 0.06 & 0.18 & 0.11 & 0.19 & 0.14 \\
        \midrule
         \multirow{2}{*}{PISCO} & \multirow{2}{*}{Unseen} & 1 per sample & 0.49 & 0.56 & 0.58 & 0.39 & 0.42 & 0.50 \\
         & & 1 per sent. & 0.60 & 0.65 & 0.71 & 0.51 & 0.49 & 0.63 \\
        \bottomrule
    \end{tabular}
    }
    \caption{\tocheck{Fraction of entities preserved in the reconstructed context using xRAG and PISCO models.}}
    \label{tab:res_xrag_entity_preservation}
\end{table}

We evaluate xRAG on two datasets: \emph{``seen''} data--random samples from xRAG's pre-training set, and \emph{``unseen''} data--samples from HotpotQA, QuAC, and TriviaQA. Each dataset contains 450 examples, split into three 150-example sets with 1, 5, and 10 sentences per example, respectively. Our primary metric is BERTScore, which can identify meaning-preserving paraphrases as accurate reconstructions.
Table~\ref{tab:res_xrag_reconstructing} indicates that the xRAG method is far from being able to reconstruct the original context (average BERTScore F1 is 0.66). On average the performance on \emph{seen} and \emph{unseen} examples is similar, suggesting that xRAG is learning a general representation rather than memorizing its pre-training data. Additionally, the target LLM is not able to handle more than one xRAG token, and the reconstruction scores drop by 20-30 BERTScore F1 points when using one token per sentence.

Higher scores for examples containing multiple sentences likely come from the fact that xRAG pre-training data contains longer samples and the model is not able to preserve more granular pieces of information (see Figure~\ref{fig:xrag_pre-training_data_distribution}).
The qualitative analysis shows that xRAG tokens primarily capture the general topic of the compressed content (see Table~\ref{tab:res_xrag_reconstruction_examples} in Appendix~\ref{app:add_results}) but fail to retain key details. This pattern is further supported by our entity preservation experiments, where we measure the fraction of text entities from the input retained in the reconstructed version (last column in Table~\ref{tab:res_xrag_reconstructing}). 

\tocheck{The fraction of entities' types preserved in the reconstructed context per entity type is presented in Table~\ref{tab:res_xrag_entity_preservation}. We observe particularly low scores for \emph{dates} and \emph{numerical} values. Additionally, entity preservation for \emph{people} is notably poor on in-domain samples, likely due to noise or ambiguous references in the pre-training data (e.g.,~``\emph{Thomas Scott 1806–1816}'' annotated as \emph{person}). In contrast, \emph{geographical locations} are the most consistently preserved across all input types. Overall, our findings indicate that the model struggles with maintaining entities during prompt compression.}

PISCO results are reported only for xRAG \emph{``unseen''} data for fair comparison and show significantly higher information preservation---both in terms of semantic similarity between the reconstructed and original text, and the percentage of preserved entities. Notably, PISCO maintains a similar level of information retention across different encoding levels, likely due to its higher encoding capacity: while xRAG compresses information into a single token, PISCO uses 8 embeddings.

\subsection{Compression Computational Efficiency}
\label{subsec:results:add_results:flops}

\tocheck{We evaluate the computational cost of various prompt compression methods by measuring the total number of floating-point operations (FLOPs) required to compress a standard input sample (context documents for a randomly selected HotpotQA dataset sample). The results, summarized in Table~\ref{tab:flops}, provide a comparison of the efficiency of each compression method, highlighting trade-offs between compression strategy and computational demand. Dominating operation remains the number of matrix multiplication. The results show that all compression methods used in our experiments are comparable in terms of efficiency.} 

\begin{table}[tp]
    \centering
    \setlength{\tabcolsep}{1pt}
    \scriptsize
    \begin{tabular}{lc}
        \toprule
        \textbf{Method} & \textbf{MFLOPs} \\
        \midrule
        xRAG & \(3.7 x 10^7\) \\
        PISCO & \(1.3 x 10^7\) \\
        LLMLingua & \(6.6 x 10^6\) \\
        \bottomrule
    \end{tabular}
    \caption{\tocheck{Computational cost of different prompt compression methods measured in floating-point operations (FLOPs). Values are reported in MFLOPs, representing the total number of FLOPs required for processing a single input sample.}}
    \label{tab:flops}
\end{table}

%% file: sections/05-results.tex
\section{Improving xRAG}
\label{sec:results}

The results presented in Section~\ref{sec:analysis_prompt_compression} show a large disproportion between PISCO and the remaining two prompt compression methods. 
In particular, we highlight xRAG's inability to preserve detailed information from the original input in long-context scenarios. 
We here explore 
directions to mitigate these limitations of xRAG, while still retaining the advantage
of high compression rates. We use our prompt compression evaluation framework to verify whether the modifications lead to improvements over the original xRAG method. 
We re-train an xRAG model using the data and code released with its paper to ensure that the models are fully comparable (details in Appendix~\ref{app:reproducibility_study}). 
We denote this reproduced version as \emph{xRAG (Reproduced)}.

\subsection{Sentence Pre-Training and Sentence-Level Fine-Tuning (\emph{w/ Sentence PT + FT)}}
\label{subsec:spt_ft}

\tocheck{The only trainable component of the xRAG is a bridge model, whose main role is to align two embedding spaces -- the one from the encoder (dense retrieval) model and the one from the target LLM. This function is learned via unsupervised pre-training on a reconstruction task, and then further fine-tuned on a set of target downstream tasks.}

\begin{figure}[t]
    \centering
    \includegraphics[width=.9\linewidth]{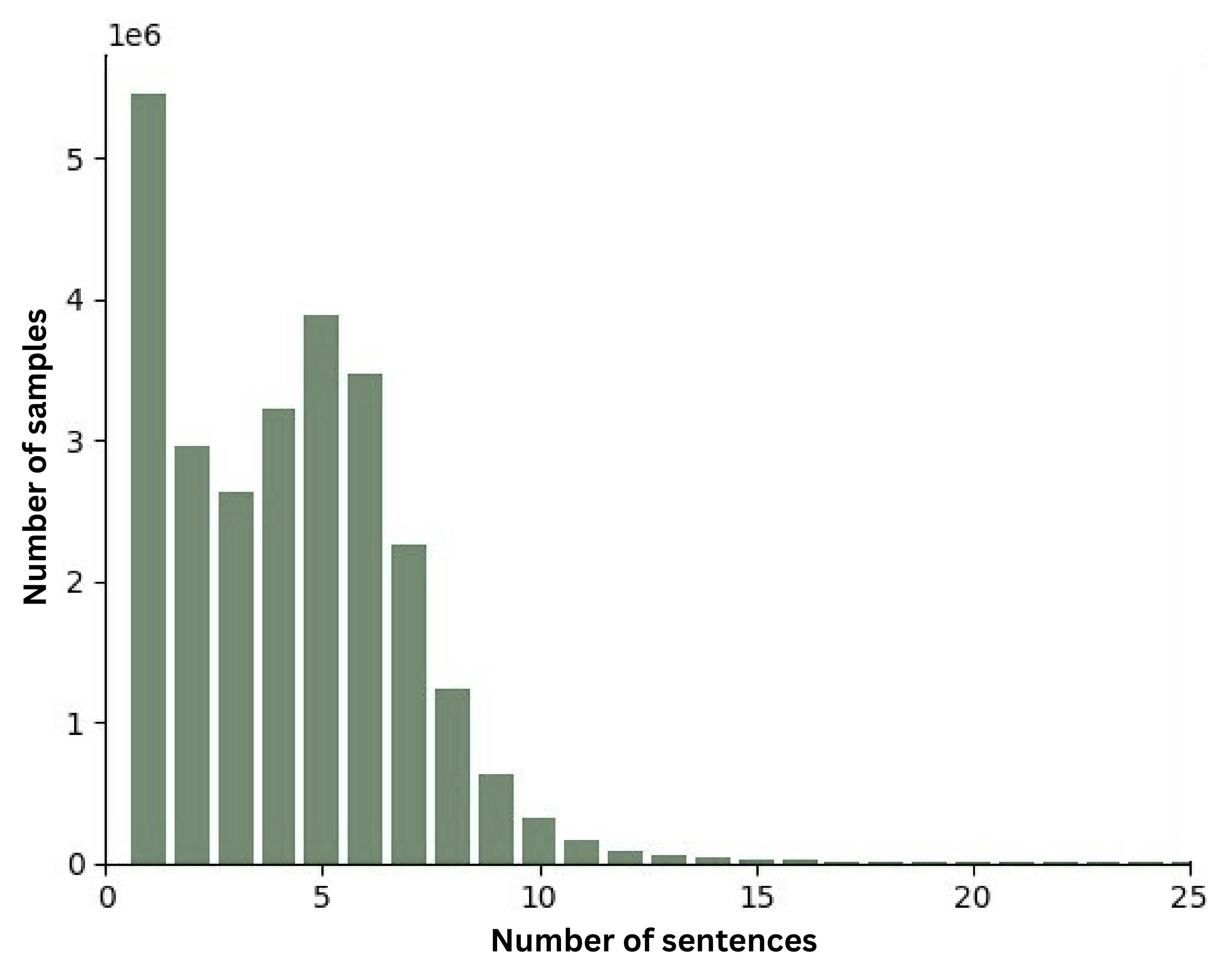}
    \caption{\tocheck{Distribution of the lengths of the examples measured in terms of number of sentences per training sample in pre-training data. The training partition contains 26,541,264 samples overall. The longest sample contains 137 sentences.}}
    \label{fig:xrag_pre-training_data_distribution}
\end{figure}

In xRAG, the pre-training (PT) samples are encoded into single xRAG tokens,
regardless of their length. Since the majority of the examples are longer than one sentence \tocheck{(see Figure~\ref{fig:xrag_pre-training_data_distribution})}, the bridge model fails to transform the representations of texts with smaller granularity (see Table~\ref{tab:res_baselines_xrag_per_tokens}) and thus to capture details beyond the topic of the samples.
To address this limitation, we use single-sentence samples in pre-training to ensure that the model can effectively handle these basic cases. We use the same pre-training data, but consider only the first sentence from each sample. In the fine-tuning (FT) phase, original xRAG chunks each context into texts of 180 tokens with each chunk encoded into one xRAG token. Instead, we replace token-based chunking with sentence-based segmentation with each sentence encoded into a separate xRAG token to preserve information continuity. We denote this variant of xRAG as \emph{xRAG w/ Sentence PT + FT}.

\begin{table}[tp]
    \centering
    \resizebox{\columnwidth}{!}{%
    \setlength{\tabcolsep}{3pt}
    \begin{tabular}{llccccc}
        \toprule
         \textbf{Method} & \textbf{xRAG} & \textbf{HotpotQA} & \textbf{HotpotQA*} & \textbf{arXiv-sum.} & \textbf{QuAC} & \textbf{TriviaQA} \\
        \midrule
        xRAG (Reproduced) & \multirow{3}{*}{1 per cont.} & 0.50 & 0.58 & 0.38 & 0.43 & 0.72 \\
         w/ Sentence PT + FT & & \textbf{0.52} & \textbf{0.62} & \textbf{0.44} & \textbf{0.49} & \textbf{0.73} \\
         w/ Two-Step PT + FT &  & 0.40 & 0.45 & 0.35 & 0.42 & 0.68 \\
        \midrule
         xRAG (Reproduced) & \multirow{3}{*}{1 per sent.} & 0.31 & 0.47 & 0.26 & \textbf{0.47} & \textbf{0.71} \\
         w/ Sentence PT + FT & & \textbf{0.34} & 0.52 & \textbf{0.46} & 0.42 & 0.65 \\
         w/ Two-Step PT + FT &  & 0.23 & \textbf{0.58} & 0.30 & 0.39 & 0.55 \\
        \bottomrule
    \end{tabular}
    }    
    \caption{Grounding scores w.r.t. the context, averaged across 5 random sets of 100 examples ($\sigma < 0.1$), calculated over the non-empty outputs (see Appendix~\ref{app:add_results:grounding}).}
    \label{tab:res_xrag_variants_grounding}
\end{table}

\begin{figure}[tp]
    \centering
    \includegraphics[width=\columnwidth]{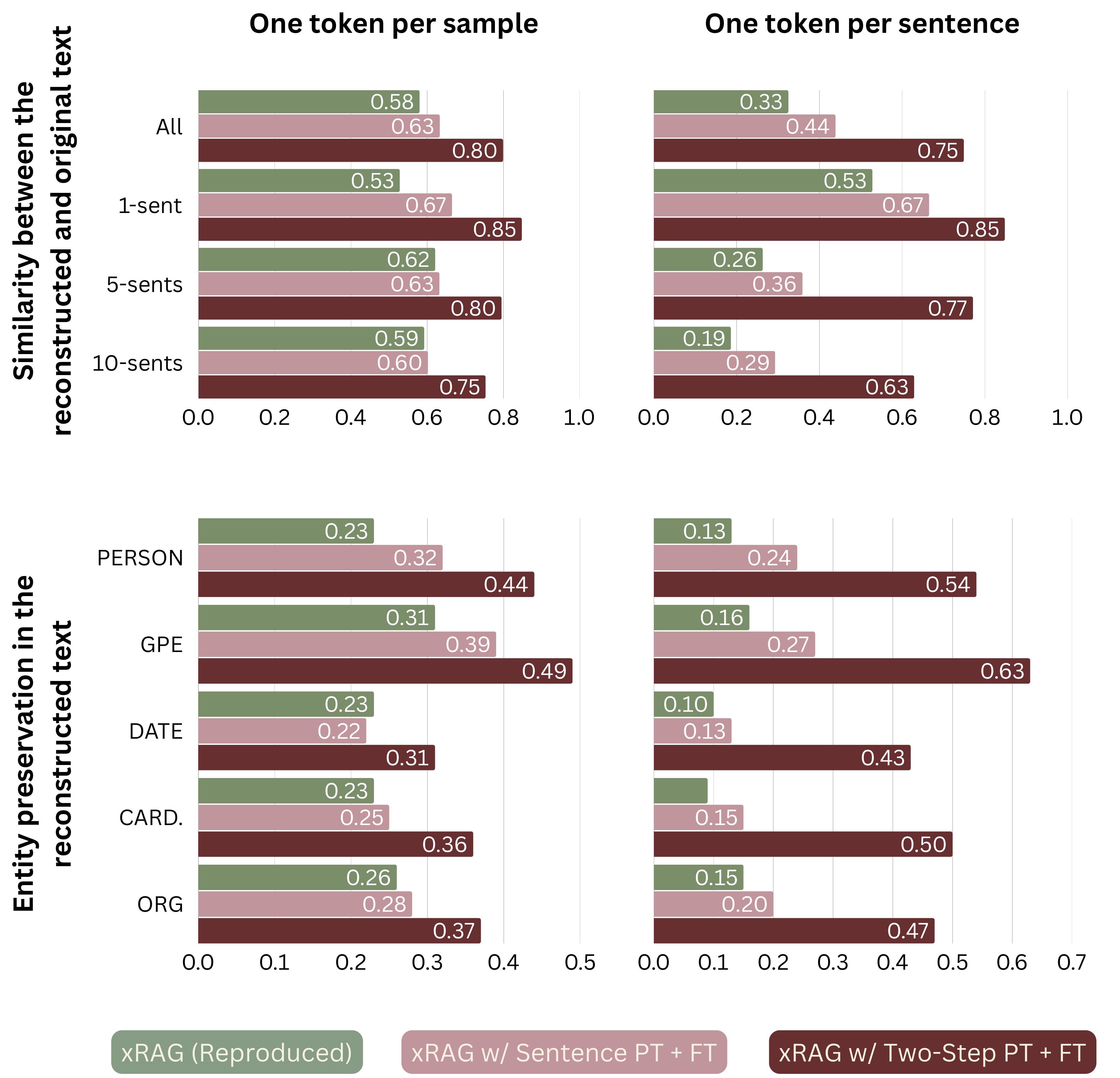}
    \caption{Information preservation results for xRAG variants. Similarity is measured with BERTScore between the original and reconstructed text. Entity preservation is based on EM of entities in the reconstruction.}
    \label{fig:res_xrag_variants_reconstruction_results}
\end{figure}

On the downstream evaluation, \emph{xRAG w/ Sentence PT + FT} brings sizable improvements on all tasks (Table~\ref{tab:res_xrag_variants_downstream_tasks}). With context-level (\emph{Cont.}) compression we see 1-5 points absolute gains on all tasks (up to 13\% relative), suggesting that this method allows to capture the information better. This is further supported by the increase in grounding of the generated text (Table~\ref{tab:res_xrag_variants_grounding}). We also see that the model is able to handle sentence-level compression better (\emph{Sent.}) -- especially on HotpotQA* (downstream performance increases from 0.174 to 0.423) and TriviaQA (from 0.115 to 0.361). Interestingly, sentence-level representations are still not performing as well as the context-level ones. 

Finally, in the context reconstruction experiments, we see improved performance when encoding at sentence level.\footnote{Reconstructed samples are in Table~\ref{tab:res_xrag_reconstruction_examples} in Appendix~\ref{app:add_results}.} The \emph{xRAG w/ Sentence PT + FT)} model is able to preserve more entities on average---increase from 13\% to 20\% (see Figure~\ref{fig:res_xrag_variants_reconstruction_results}). However, in absolute terms, the numbers still remain low (preserving between 13\%-39\% entities). In all three dimensions, the performance on sentence-level tokens is lower than when encoding the entire context at once, implying that the model is not able to handle multiple tokens at a time.

\subsection{Two-Step~Pre-Training~and~Sentence-Level Fine-Tuning (\emph{w/ Two-Step PT + FT})}

Even though we see improvements in downstream performance and grounding with the \emph{xRAG w/ Sentence PT + FT} model (\S~\ref{subsec:spt_ft}), the sentence-level encoding is still falling behind. We hypothesize that the main reason is that the model is not able to handle information spread across multiple tokens. 

To verify and refine this hypothesis, we introduce an additional pre-training step that focuses on helping the model reason about information from multiple xRAG tokens. We adopt a two-stage pre-training procedure with these steps: (\emph{i})~encoding one sentence per pre-training sample; (\emph{ii})~chunking samples into sentences with each sentence encoded separately. The fine-tuning stage is the same as in \emph{xRAG w/ Sentence PT + FT} (chunking samples into sentences with one token per sentence). We denote this variant as \emph{xRAG w/ Two-Step PT + FT}.

Results on downstream tasks show that prompt compression methods can further benefit from sentence-level encoding, particularly for datasets with focused, concise contexts like HotpotQA*, where low-level facts are needed for reasoning (Table~\ref{tab:res_xrag_variants_downstream_tasks}). We see 23\% relative improvement over the baseline (in EM, from 0.38 for \emph{xRAG (Reproduced) Cont.} to 0.47 for \emph{xRAG w/ Two-Step PT + FT Sent.}).
We generally observe consistent improvements over the baseline with this model when the information is encoded at sentence-level (\emph{Sent.} columns).
Still, the model cannot fully process complex, noisy contexts, as evident in results for HotpotQA and TriviaQA, where the performance with one token per context drops (\emph{Cont.} columns). 

Even though we observe lower grounding for multi-token encoded texts, the reconstruction experiments show that our two-step pre-training on data with different granularity results in significant improvements in terms of information preservation. Crucially, this effect is
visible independently of the amount of encoding tokens used. 
On average, we can reconstruct 50\% of entities from the original text encoded into multiple tokens (compared to 13\% for the \emph{xRAG (Reproduced)}) (Figure~\ref{fig:res_xrag_variants_reconstruction_results}). 
While cardinal entities remain the most challenging ones to preserve, the two-step model retains significantly more of them (0.36 vs.~0.23 at  context-level and 0.50 vs.~0.09 at sentence-level). Moreover, we observe even higher similarity scores for text encoded into multiple tokens (F1 of 0.63 vs.~0.19), 
showing that our strategy captures finer-grained information.

\subsection{Qualitative Analysis}

\tocheck{To understand the type of information preserved, we conducted a qualitative analysis of reconstructed examples, examining 10 samples of varying lengths generated using 20 different prompts. Our investigation focused on prompts from the xRAG paraphrase training phase. The analysis confirmed our hypothesis that the xRAG tokens primarily capture the general topic of the compressed content but fail to retain specific details. This behavior is expected, as xRAG relies on a dense retrieval model trained for document similarity. Depending on the topic, the LLM can reconstruct the main entity described in the text, aligning with its strong performance on the TriviaQA dataset. However, numerical entities such as dates and numbers are often lost. Examples of reconstructed contexts can be found in Table~\ref{tab:res_xrag_reconstruction_examples} in Appendix~\ref{app:add_results}.}

\subsection{Compression Rates}
Table~\ref{tab:res_baselines_prompt_lengths} presents the compression rates for each model (measured in tokens and including the system instructions, context, etc.).
PISCO achieves very high compression rates without performance loss in any of the aspects investigated with our evaluation framework, establishing it as the state-of-the-art for prompt compression in long-context scenarios.
The sentence-level xRAG retains a high compression rate with a 4x-12x reduction in the input size and a 2x higher compression rate than LLMLingua. The compression rate depends on the lengths of the inputs but the compression granularity is a controllable parameter of the method.

\begin{table}[tp]
    \centering
    \resizebox{\columnwidth}{!}{%
    \setlength{\tabcolsep}{3pt}
    \begin{tabular}{lcccccc}
        \toprule
        \textbf{Method} & \textbf{HotpotQA} & \textbf{HotpotQA*} & \textbf{arXiv-sum.} & \textbf{QuAC} & \textbf{TriviaQA} & \textbf{GSM8K} \\ %
        \midrule
        Mistral-7B & 1515 & 341 & 6424 & 995  & 840 & 1135 \\ %
        Mistral-7B (no cont.) & 44 & 44 & --- & 162 & 40 & 109 \\ %
        xRAG (1 per cont.) & 65 (23.3) & 65 (5.2) & 27 (237.9) & 186 (5.3) & 61 (13.8) & 130 (8.7) \\ %
        xRAG (1 per sent.) & 148 (10.2) & 80 (4.3) & 525 (12.2) & 234 (4.3) & 103 (8.2) & 172 (6.6) \\ %
        PISCO & 73 (20.8) & 73 (4.7) & 35 (183.5) & 194 (5.1) & 69 (12.2) & 138 (8.2) \\ %
        LLMLingua & 388 (3.9) & 312 (1.1) & 329 (19.5) & 398 (2.5) & 362 (2.3) & 417 (2.7) \\ %
        \bottomrule
    \end{tabular}
    }
    \caption{Average prompt lengths (in tokens) after compression. The compression rate is shown in brackets. See Table~\ref{tab:flops} for compression computational efficiency.}
    \label{tab:res_baselines_prompt_lengths}
\end{table}

%% file: sections/06-conclusions.tex
\section{Lessons Learned and Future Directions}
\label{sec:lessons}

So far, we have seen that many compression techniques fail to preserve detailed information, especially in long-context scenarios and at high compression rates, leading to drops in performance and grounding.
In line with this, context reconstruction experiments indicate that xRAG tokens primarily capture the general topic of the compressed content but fail to retain essential details for a task~(numbers, dates, names, etc.). 
We propose modifications of xRAG and provide evidence that manipulating information granularity during training improves information retention in soft prompts, which translates to improved performance and grounding in downstream tasks.
Notably, improvements occur regardless of the number of encoding tokens used.

\paragraph{Multi-Token Context Understanding}
Given the complexity and information density of text in information-intensive tasks, an effective alternative to encoding everything into a single soft prompt token is needed~\citep{Mu:2023:NIPS, Li:2024:arXiva}. 
\tocheck{Our experiments suggest that while increasing text granularity is promising, reasoning across information stored independently in soft prompt tokens remains challenging. PISCO addresses this by using sequence-level distillation, rather than token-level~\citep{louis2025pisco}. Our results confirm that the training with multi-token samples implemented in PISCO significantly improves grounding and information preservation and is crucial for the ability to integrate information across separate units~\citep{Cheng:2024:NIPS}.} 

\paragraph{Context-Aware Adaptive Compression}
Multi-token soft prompting is an important direction for compression methods. However, a task-independent \emph{one-size-fits-all} strategy does not capture the fact that every input comes with different complexity and level of detail required for the target task~\cite{Chevalier:2023:EMNLP,Mu:2023:NIPS,Jiang:2023:EMNLP,pan-etal-2024-llmlingua}. Moreover, despite the advancements in prompt compression methods, grounding and information retention remain suboptimal, especially when encoding with multiple tokens. Dynamically adjusting the compression rate and the compression strategy can improve contextual understanding on information spread in several tokens~\citep{nagle2024fundamental}.

\paragraph{Task-Specific Embeddings}
One of the main advantages of xRAG-like methods is the fact that context encoding can be done with a different model than the target LLM. This model can be smaller and more efficient, or larger and more capable than the target LLM.
xRAG's reliance on a dense retrieval model trained for document similarity explains its tendency to only capture high-level topic information while overlooking specific details. A promising direction is exploring alternative models, such as instruction-tuned embeddings~\citep{su-etal-2023-one,behnamghader2024llmvec,wang2024multilingual,wang-etal-2024-improving-text}, which can provide task-specific representations~(conditioned on a set of instructions) and could better preserve entity-level information.

\paragraph{Hybrid Soft-Hard Prompting}
Preserving details, such as named entities, in compressed representations is crucial for improving performance on tasks that require reasoning. 
Hard compression~\cite{li-etal-2023-compressing,Jiang:2023:EMNLP,chuang-etal-2024-learning} is inherently explainable and can explicitly retain these details. In contrast, soft prompting has limited representational capacity, making it challenging to encode certain information, such as cardinals, in highly compressed embeddings~\cite{wallace-etal-2019-nlp,10.1162/tacl_a_00324,epure-hennequin-2022-probing}. 
A promising approach combines both: hard prompts preserve key details (e.g., numbers, names), while the rest of the context is encoded via soft prompts, reducing information loss during the compression.

\section{Conclusions}
In this paper, we propose a framework for assessing the quality of prompt compression methods on long-context generation tasks. We shed light on both the capabilities and limitations of the state-of-the-art hard~(LLMLingua) and soft~(xRAG and PISCO) compression approaches. Our findings reveal that some existing compression techniques fail to preserve detailed information, especially in long-context scenarios. We explore several directions to improve the xRAG soft compression method, showing an increase in model performance -- up to +23\% on downstream tasks, up to 8 BERTScore points in grounding, and preserving 2.7x more entities.\footnote{The code repository: \url{https://github.com/amazon-science/information-preservation-in-prompt-compression}.}

\section*{Limitations}

Our study has several limitations that should be considered when interpreting the results. First, all experiments are conducted using Mistral-7B as the target LLM. While this allows for consistent evaluation, exploring different model sizes and architectures is beyond the scope of this work and remains an area for future research.
Second, our analysis focuses exclusively on English benchmarks. Findings may not generalize to multilingual settings, where language-specific characteristics could influence the effectiveness of prompt compression methods. \tocheck{We believe that our findings will not change across languages, as we outline fundamental issues with the compression frameworks. If anything, we expect the soft approaches to be even more shallow in capturing details in other languages.}
Third, we evaluate only one representative baseline from each category of prompt compression techniques. While this provides an initial comparison, a broader investigation using our evaluation framework is needed to fully understand the trade-offs across different methods.
Finally, although we examine multiple generation tasks, there is still room for further exploration, particularly in long-form QA scenarios, where prompt compression may have a more significant impact on model performance.

%% file: sections/07-appendix.tex
\section*{Appendix for paper ``\PaperTitle''}

\section{Implementation Details}

\paragraph{Models}
In all the experiments we use \texttt{Mistral-7B Instruct v0.2}~\citep{Jiang:2023:arXiv}\footnote{\href{https://huggingface.co/mistralai/Mistral-7B-Instruct-v0.2}{huggingface.co/mistralai/Mistral-7B-Instruct-v0.2}} that contains approximately 7.24B parameters as a target LLM. We use the associated model checkpoint implemented using the PyTorch~\citep{NEURIPS2019_bdbca288} framework that is available on the HuggingFace Hub via the transformers~\citep{wolf-etal-2020-transformers} library. We access \texttt{microsoft/phi-2}~\citep{javaheripi2023phi}\footnote{\url{https://huggingface.co/microsoft/phi-2}} for LLMLingua and the Mistral 7B based xRAG (\texttt{Hannibal046/xrag-7b}\footnote{\url{https://huggingface.co/Hannibal046/xrag-7b}\label{xrag:model}}) in the same way. We load the datasets from the HuggingFace using the Datasets library~\cite{lhoest-etal-2021-datasets}.

\paragraph{Training}
We train our models using two compute instances, each equipped with 8 NVIDIA A100 Tensor Core GPUs, 1152 GiB of memory, and 96 vCPUs. The training process for a new xRAG model variant typically takes around 16 hours. We adopt a batch size of 1 per GPU and leverage the Accelerate library with a gradient accumulation step of 4.\footnote{\url{https://huggingface.co/docs/accelerate}} The specific hyper-parameters used during training are detailed in Table~\ref{tab:hyperparameters}.

The tuning dataset for xRAG models exhibits some inconsistencies. The combined datasets contain a total of 1,128,075 samples. However, only 628,667 samples include a context component, and after filtering out instances where the background text is shorter than three words, 628,003 samples remain. The remaining 499,408 samples originate from QA or fact-checking datasets that do not provide explicit context.

\begin{table}[tp]
    \centering
    \resizebox{\columnwidth}{!}{%
    \setlength{\tabcolsep}{3pt}
    \begin{tabular}{lcc}
        \toprule
         \textbf{Hyperparameter} & \textbf{Pre-training} & \textbf{Fine-tuning} \\
        \midrule
        Optimizer & AdamW & AdamW \\
        Learning rate & 6e-3 & 2e-5 \\
        LR scheduler type & linear & linear \\
        Warmup ratio & 0.03 & 0.03 \\
        Weight dacay & 0.0 & 0.0 \\
        Epochs & 1 & 1 \\
        KL \(\alpha\) & --- & 2.0 \\
        KL temperature & --- & 1.0 \\
        Flash attention & True & True \\
        Batch size & 1 & 1 \\
        Gradient accumulation steps & 4 & 2 \\
        Num GPUs & 8 & 8 \\
        \bottomrule
    \end{tabular}
    }
    \caption{Hyper-parameters used during the pre-training and fine-tuning phases of the xRAG model and its variants.}
    \label{tab:hyperparameters}
\end{table}

\paragraph{Evaluation}

Evaluation is conducted on a subset of the available datasets due to the high number of datasets and model combinations. To ensure consistency, samples are randomly selected using a fixed seed of 42. Response generation follows a greedy search decoding strategy, while context truncation varies depending on the model. For xRAG, the retriever tokenizer processes an input truncated to 16,000 tokens, but no truncation is applied to the prompt passed to the target language model. Mistral, on the other hand, limits input to 8,192 tokens, a necessary constraint given that an Arxiv article tokenized with Mistral's tokenizer can reach 35,208 tokens, leading to out-of-memory (OOM) errors if not truncated. LLMLingua processes the full input for compression and then truncates the compressed prompt to 8,000 tokens when required before generation.
For generation settings, the target token count in LLMLingua is fixed at 350, aligning with the best-reported performance in prior work. Mistral employs a \emph{max\_new\_tokens} setting of 500. 
The evaluation metric in the LLMLingua reproducibility study uses human-written summaries from the dataset as ground-truth references, allowing for an assessment of the generated responses against high-quality, manually curated summaries.

\paragraph{Data Processing}

We use the validation partition from the "rc" TriviaQA subset, retaining 9,533 samples after removing those without context. Each sample includes either Web search results or Wikipedia articles, but we use only the first Web search result to better simulate a realistic retrieval-augmented generation (RAG) scenario. To manage lengthy documents, which range from 1 to 8,246 sentences (averaging 124), we limit the context to a maximum of 50 sentences, resulting in 5,743 samples, while 7,479 samples remain within 100 sentences. For evaluation, we compute results on 863 out of 1,000 QuAC validation samples, discarding those with fewer than four question-answer pairs, as the fourth question serves as the model's prompt. Additionally, in-context learning (ICL) demonstrations for GSM8K are sampled from its training partition.

\paragraph{Reproducibility of our Experiments}

To ensure experiment reproducibility, we use a fixed pseudo-random seed when sampling evaluation examples, allowing for consistent dataset selection across runs. Additionally, we set the model's temperature to 0 during generation, ensuring a fully deterministic response process. These measures eliminate variability in both data sampling and model outputs, enabling reliable comparisons and replication of results.

\section{Experimental Setup}
\label{app:experimental_setup}

\subsection{Evaluation Metrics}
\label{app:experimental_setup:metrics}
We use the standard similarity metrics for evaluating the downstream performance: (\emph{i})~for summarization and long-form QA we use ROUGE~\citep{Lin:2004:ACL}\footnote{\url{https://huggingface.co/spaces/evaluate-metric/rouge}} 
and BERTScore~\citep{Zhang:2020:ICLR};\footnote{\url{https://huggingface.co/spaces/evaluate-metric/bertscore}} (\emph{ii})~for short-form QA and mathematical reasoning, we use Exact Match (EM) with normalization of the generated outputs (splitting text into alpha-numeric tokens and non-whitespace tokens, ignoring capitalization and multi-lines).

\subsection{Grounding Evaluation}
\label{app:experimental_setup:grounding}

As part of our preliminary analysis, we evaluated several grounding approaches. Our analysis showed that methods based on Natural Language Inference (NLI) models -- classifying text as either entailed or not in the source -- perform poorly when dealing with long source texts and answers requiring information synthesis or reasoning. Similarly, using LLM-as-a-Judge through the RAGAs framework~\citep{Es:2024:EACL} yielded factual correctness scores skewed towards 0, making it difficult to establish a reasonable threshold. More promising results were obtained by directly prompting an LLM (\texttt{Claude 3 Haiku})\footnote{\url{https://www.anthropic.com/claude/haiku}} to score the grounding of sentences from generated responses in source paragraphs, particularly for summarization tasks. However, these generated grounding scores tended to skew towards higher values.
Given the lack of sensitivity of these approaches to grounding evaluation in our long-context scenarios, we adopted a state-of-the-art approach for automatic faithfulness evaluation (FABLES,~\citet{Kim:2024:CML}), as it shows balanced scores and is able to handle long contexts.

\section{Datasets}
\label{app:datasets}

\subsection{Training Dataset}
\label{app:datasets:training}

Figure~\ref{fig:xrag_pre-training_data_distribution} presents the distribution of a number of sentences per training sample in xRAG pre-training data. The dataset contains more than 26M Wikipedia snippets from \href{https://github.com/facebookresearch/atlas?tab=readme-ov-file#models}{enwiki-dec2021}.

\begin{figure*}[t]
\centering
\subfigure[HotpotQA]{\includegraphics[width=0.30\linewidth]{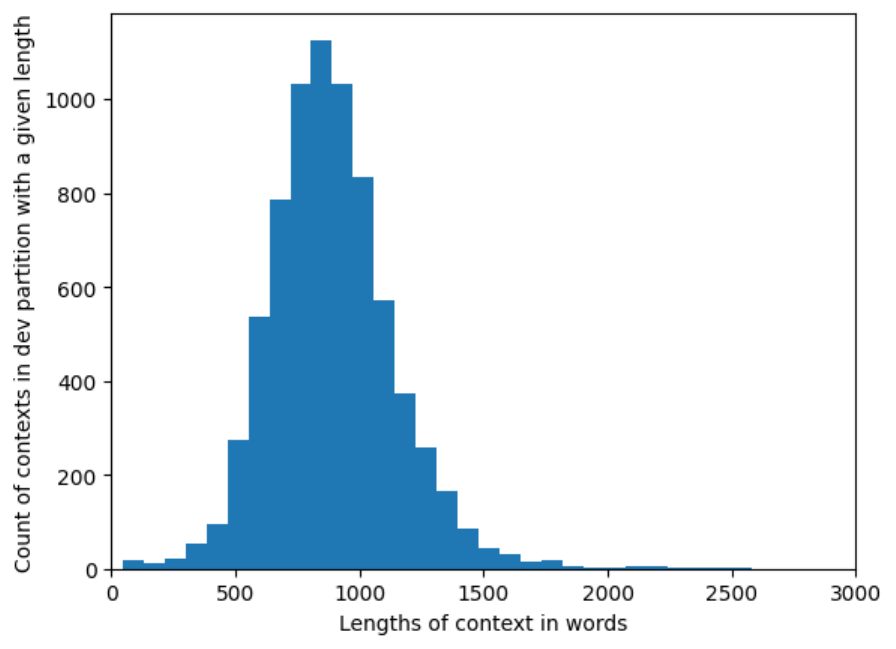}}
\subfigure[arXiv-summarization]{\includegraphics[width=0.30\linewidth]{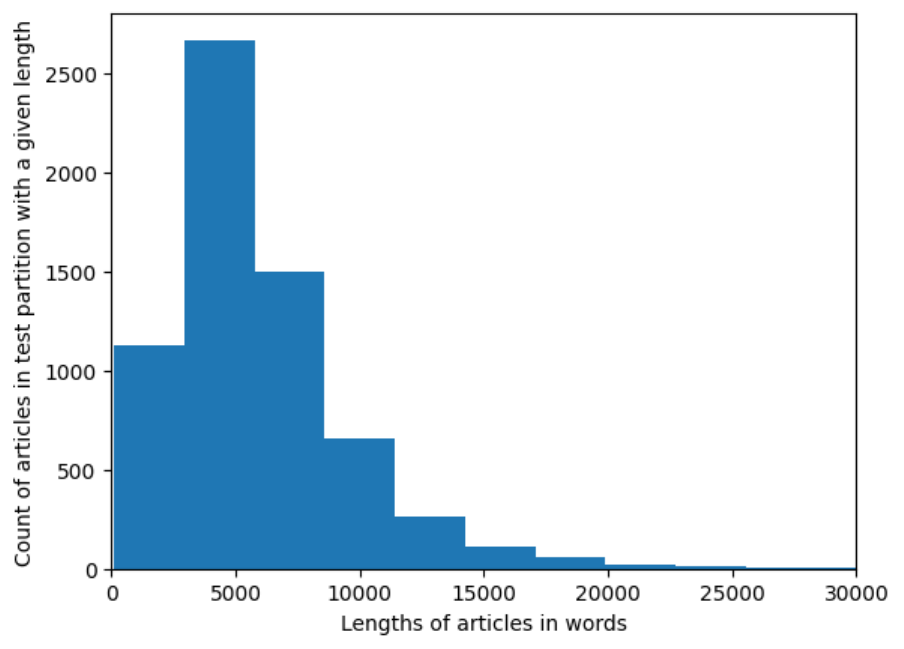}}
\subfigure[QuAC]{\includegraphics[width=0.30\linewidth]{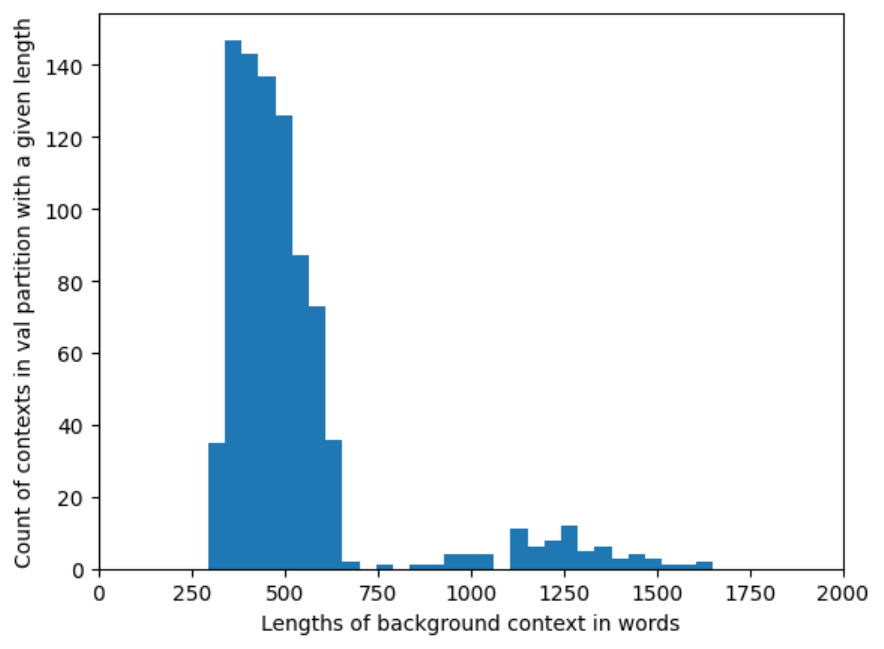}}
\subfigure[TriviaQA]{\includegraphics[width=0.30\linewidth]{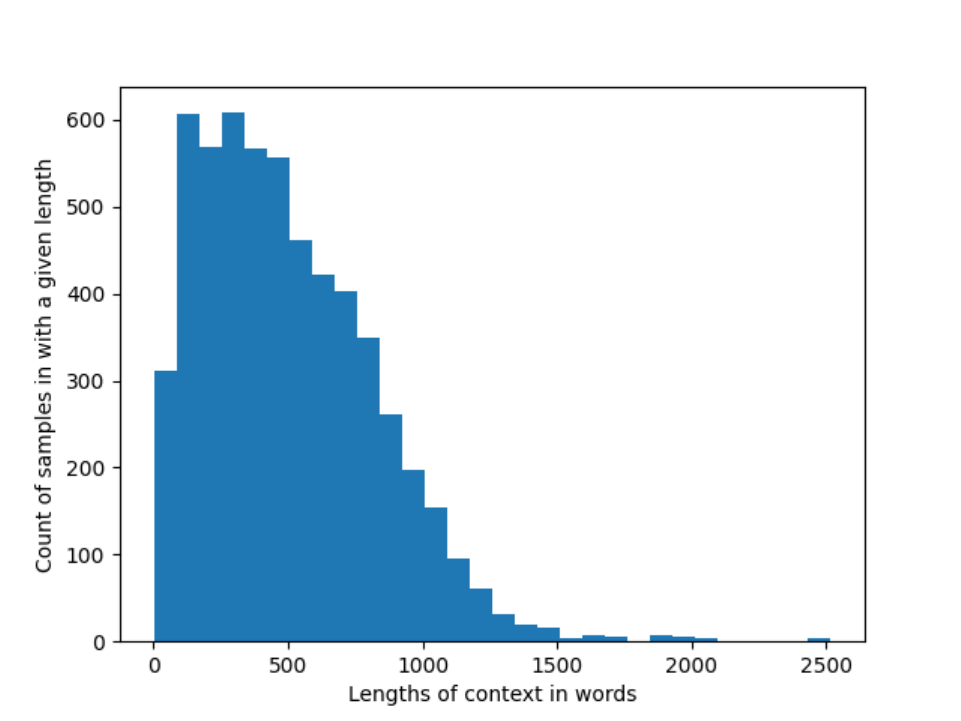}}
\subfigure[GSM8K]{\includegraphics[width=0.30\linewidth]{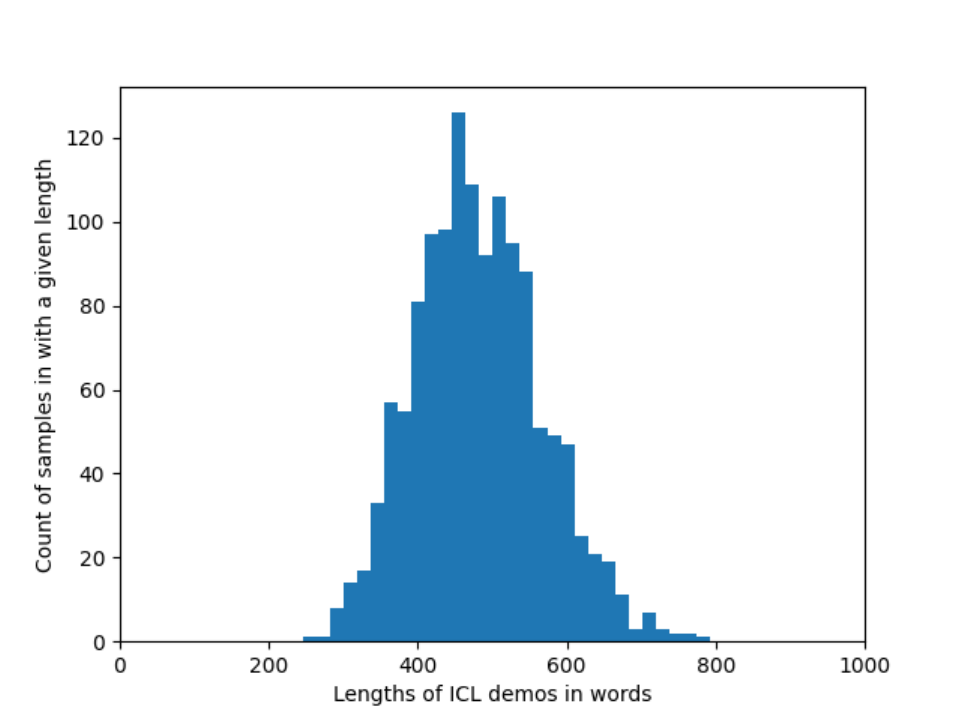}}
\caption{Histograms of context lengths in the different datasets we use for evaluation.}
\label{fig:datasets_context_length_distribution}
\end{figure*}

\subsection{Evaluation Datasets}
\label{app:datasets:eval}

\paragraph{HotpotQA} is a question-answering dataset featuring natural, multi-hop questions. It requires reasoning over supporting facts and enables more explainable question-answering systems. It contains three types of questions: 
\begin{enumerate}[nosep,leftmargin=*,label=(\roman*)]
    \item single-hop questions that require reasoning over one of the paragraphs
    \item multi-hop questions that were correctly answered by the model but require reasoning over multiple documents
    \item hard multi-hop questions that were not correctly answered by the model and require reasoning over multiple documents
\end{enumerate}
We generate answers for 1,000 examples randomly sampled from the development partition and use them for evaluation.

\paragraph{Arxiv-sumarization} is a long document summarization dataset. The test partition used in our experiments contains 6,440 arXiv articles divided into sections. We treat the abstracts of the articles as a proxy for a human-written ground-truth summary. 

\paragraph{Question Answering in Context (\emph{QuAC})} is a dataset for modeling, understanding, and participating in information-seeking dialog. Data samples are in the form of interactive dialogs between two crowd workers: (\emph{i})~a student who poses a sequence of free-form questions to learn as much as possible about a given topic, and (\emph{ii})~a teacher who answers the questions by providing short spans from the text of the Wikipedia text that is hidden from the student. 
Results are computed for 863 out of 1,000 samples from the QuAC validation partition. We discard samples that have fewer than 4 question-answer pairs (turns in the conversation). The fourth question is the question we prompt the model to answer. 

\paragraph{Grade School Math 8K (\emph{GSM8K})} is a dataset of 8.5K high-quality, linguistically diverse grade school math word problems, designed for evaluating question-answering models on basic mathematical tasks. Solving these problems requires multi-step reasoning, typically involving 2 to 8 steps of elementary arithmetic operations. The questions do not require concepts beyond early Algebra, and most can be solved without explicitly defining variables. Solutions are presented in natural language rather than pure mathematical expressions.

\paragraph{TriviaQA} is a large-scale reading comprehension dataset containing over 650K question-answer-evidence triples, including 95K expert-authored question-answer pairs with independently collected evidence documents. The dataset features complex, compositional questions with high syntactic and lexical variability which require cross-sentence reasoning. This makes TriviaQA more challenging than other large-scale QA datasets.

\begin{table*}[tp]
    \centering
    \resizebox{0.9\textwidth}{!}{%
    \setlength{\tabcolsep}{3pt}
    \begin{tabular}{llcc}
        \toprule
        \multirow{2}{*}{\textbf{Method}} & \multirow{2}{*}{\textbf{Context}} & \multicolumn{2}{c}{\textbf{Exact match}} \\
         & & \textbf{dev\_distractor\_v1 (1000)} & \textbf{test-distractor} \\
        \midrule
        \multirow{2}{*}{Mistral-7B (no compression)} & All 10 context paragraphs & 0.713 & --- \\
         & Only paragraphs with supporting facts & 0.787 & --- \\
        \midrule
        \multirow{2}{*}{xRAG (Vanilla)} & All 10 context paragraphs & 0.286 & --- \\
         & Only paragraphs with supporting facts & 0.382 & --- \\
        \multirow{2}{*}{xRAG (Ours, Reproduced)} & All 10 context paragraphs & 0.286 & \\
         & Only paragraphs with supporting facts & 0.375 & --- \\
        xRAG (Paper, \citet{Cheng:2024:NIPS}) &  One retrieved document & --- & 0.340 \\
        \bottomrule
    \end{tabular}
    }
    \caption{Reproducibility results for the xRAG approach. This table compares different xRAG variants: the \emph{vanilla} version, based on the publicly available checkpoint (without modification); the \emph{reproduced} version, which is our re-trained model using the original codebase (Appendix~\ref{app:reproducibility_study}); and xRAG (\emph{Paper}), presenting the results reported in the xRAG paper.}
    \label{tab:reproducibility_xrag}
\end{table*}

\begin{table*}[tp]
    \centering
    \resizebox{0.9\textwidth}{!}{%
    \setlength{\tabcolsep}{3pt}
    \begin{tabular}{llccccc}
        \toprule
        \textbf{Method} & \textbf{Test samples} & \textbf{BLEU} & \textbf{ROUGE-1} & \textbf{ROUGE2} & \textbf{ROUGE-L} & \textbf{BERTScore F1} \\
        \midrule
        LLMLingua (350 tokens) with Mistral-7B (Ours) & 100 random & 0.164 & 0.492 & 0.211 & 0.316 & 0.881 \\
        LLMLingua~(Paper, \citet{Jiang:2023:EMNLP}) & all (?) & 0.232 & 0.542 & 0.327 & 0.427 & 0.903 \\
        \bottomrule
    \end{tabular}
    }
    \caption{Reproducibility results of the LLMLingua approach. We compare the numbers reported in the LLMLingua paper~\citep{Jiang:2023:EMNLP} with the numbers that we obtain using the code provided by the authors (without modifications).}.
    \label{tab:reproducibility_llmlingua}
\end{table*}

\section{Reproducibility of Baselines}
\label{app:reproducibility_study}

To ensure the reliability of our baseline implementations, we conduct a reproducibility study on xRAG and LLMLingua, evaluating them on the datasets used in their respective papers. Our goal is to verify that our implementations produce comparable results.

For xRAG, we assess performance on 1,000 randomly selected samples from the HotpotQA development set (see Table~\ref{tab:reproducibility_xrag}). We evaluate two versions: (\emph{i})~the vanilla xRAG model released by the authors,\footref{xrag:model} and (\emph{ii})~our reproduced version trained using the original data and codebase (\href{https://github.com/Hannibal046/xRAG}{github.com/Hannibal046/xRAG}). The reproduced model follows the same pre-training and fine-tuning setup, though we use approximately 50\% of the fine-tuning samples with original context paragraphs instead of retrieved ones due to limitations in reproducing the original data pre-processing steps.

Our implementation of xRAG achieves slightly better results than those reported in the paper when using curated context and filtering out irrelevant, yet topically related, paragraphs. The performance gap between the vanilla and the reproduced xRAG is minimal. Additionally, Mistral-7B without prompt compression achieves results comparable to the best system on the HotpotQA leaderboard (EM of 0.775)\footnote{\url{https://paperswithcode.com/sota/question-answering-on-hotpotqa}}, confirming the competitiveness of our baseline.

We use the code provided by LLMLingua's authors\footnote{\url{https://github.com/microsoft/LLMLingua}} and verify our implementation on the \href{https://huggingface.co/datasets/liyucheng/arxiv-march-2023}{Arxiv-march-2023} dataset. Note, that we use the vanilla version of the dataset, which slightly differs from the one used in the LLMLingua paper, as the authors applied additional pre-processing steps to filter some examples. However, the paper lacks sufficient details for full reproducibility. Our results are close to, but slightly lower than, those reported in the original paper (see Table~\ref{tab:reproducibility_llmlingua}).

\begin{table*}[tp]
    \centering
    \resizebox{0.95\textwidth}{!}{%
    \begin{tabular}{lcccccccccccc}
        \toprule
        \multirow{2}{*}{\textbf{xRAG variant}} & \multicolumn{3}{c}{\textbf{HotpotQA}} & \multicolumn{3}{c}{\textbf{HotpotQA*}} & \multicolumn{3}{c}{\textbf{TriviaQA}} & \multicolumn{3}{c}{\textbf{QuAC} }\\
         & \textbf{Cont.} & \textbf{Par.} & \textbf{Sent.} & \textbf{Cont.} & \textbf{Par.} & \textbf{Sent.} & \textbf{Cont.} & \textbf{Par.} & \textbf{Sent.} & \textbf{Cont.} & \textbf{Par.} & \textbf{Sent.} \\
        \midrule
        xRAG (Reproduced) & 0.286 & 0.139 & 0.014 & 0.375 & 0.390 & 0.174 & 0.696 & --- & 0.115 & 0.829 & --- & 0.843 \\
         w/ PT + FT & 0.248 & 0.188 & 0.158 & 0.362 & 0.376 & 0.372 & 0.680 & --- & 0.557 & 0.842 & --- & \textbf{0.861} \\
         w/ Sentence PT + FT & \textbf{0.313} & 0.197 & 0.066 & 0.422 & 0.423 & 0.423 & \textbf{0.712} & --- & 0.361 & 0.833 & --- & 0.855 \\
         w/ Two-Step PT + FT & 0.277 & 0.109 & 0.145 & 0.406 & 0.428 & \textbf{0.462} & 0.685 & --- & 0.521 & 0.834 & --- & 0.823 \\
        \bottomrule
    \end{tabular}
    }
    \caption{Extended version of Table~\ref{tab:res_xrag_variants_downstream_tasks} with paragraph-level (\emph{Par.}) scores. Performance of the different variants of the xRAG method on long-context datasets. Results are provided for context encoded into a different number of xRAG tokens. HotpotQA* indicates the scenario with only supporting documents taken as context.}
    \label{tab:res_xrag_variants_downstream_tasks_extended}
\end{table*}

\begin{table*}[tp]
    \centering
    \resizebox{\textwidth}{!}{%
    \setlength{\tabcolsep}{3pt}
    \begin{tabular}{lcccccccccccccccccc}
        \toprule
        \multirow{2}{*}{\textbf{xRAG variant}} & \multicolumn{2}{c}{\textbf{HotpotQA}} & \multicolumn{2}{c}{\textbf{HotpotQA*}} & \multicolumn{2}{c}{\textbf{arXiv-sum.}} & \multicolumn{2}{c}{\textbf{TriviaQA}} & \multicolumn{2}{c}{\textbf{QuAC}} & \multicolumn{2}{c}{\textbf{GSM8K}} \\
         & \textbf{Cont.} & \textbf{Sent.} & \textbf{Cont.} & \textbf{Sent.} & \textbf{Cont.} & \textbf{Sent.} & \textbf{Cont.} & \textbf{Sent.} & \textbf{Cont.} & \textbf{Sent.} & \textbf{Cont.} & \textbf{Sent.}  \\
        \midrule
        xRAG (Reproduced) & 0.286 & 0.014 & 0.375 & 0.174 & 0.775 & 0.760 & 0.696 & 0.115 & 0.829 & 0.843 & 0.294 & 0.001 \\
        w/ PT + FT & 0.248 \textcolor{red}{(-13\%)} & 0.158 \textcolor{teal}{(1029\%)} & 0.362 \textcolor{red}{(-3\%)} & 0.372 \textcolor{teal}{(114\%)} & \textbf{0.803 \textcolor{teal}{(4\%)}} & 0.758 \textcolor{red}{(-0\%)} & 0.680 \textcolor{red}{(-2\%)} & 0.557 \textcolor{teal}{(384\%)} & 0.842 \textcolor{teal}{(2\%)} & \textbf{0.861 \textcolor{teal}{(2\%)}} & 0.207 \textcolor{red}{(-30\%)} & 0.051 \textcolor{teal}{(50\%)} \\
         w/ Sentence PT + FT & \textbf{0.313 \textcolor{teal}{(9\%)}} & 0.066 \textcolor{teal}{(371\%)} & 0.422 \textcolor{teal}{(13\%)} & 0.423 \textcolor{teal}{(143\%)} & \textbf{0.803 \textcolor{teal}{(3\%)}} & 0.739 \textcolor{red}{(-3\%)} & \textbf{0.712 \textcolor{teal}{(2\%)}} & 0.361 \textcolor{teal}{(214\%)} & 0.833 \textcolor{teal}{(0\%)} & 0.855 \textcolor{teal}{(1\%)} & 0.231 \textcolor{red}{(-21\%)} & 0.030 \textcolor{teal}{(29\%)} \\
         w/ Two-Step PT + FT & 0.277 \textcolor{red}{(-3\%)} & 0.145 \textcolor{teal}{(936\%)} & 0.406 \textcolor{teal}{(8\%)} & \textbf{0.462 \textcolor{teal}{(166\%)}} & 0.785 \textcolor{teal}{(1\%)} & 0.696 \textcolor{red}{(-8\%)} & 0.685 \textcolor{red}{(-2\%)} & 0.521 \textcolor{teal}{(353\%)} & 0.834 \textcolor{teal}{(1\%)} & 0.823 \textcolor{red}{(-2\%)} & 0.111 \textcolor{red}{(-62\%)} & 0.012 \textcolor{teal}{(11\%)} \\
        \bottomrule
    \end{tabular}
    }
    \caption{Performance of the different variants of the xRAG method. Results are provided for context encoded into a different number of xRAG tokens. Relative \textcolor{teal}{improvement}/\textcolor{red}{drop} is calculated with respect to the reproduced xRAG.}
    \label{tab:res_xrag_variants_downstream_tasks_gsm8k}
\end{table*}

\begin{figure*}[th!]
    \centering
    \includegraphics[width=\textwidth]{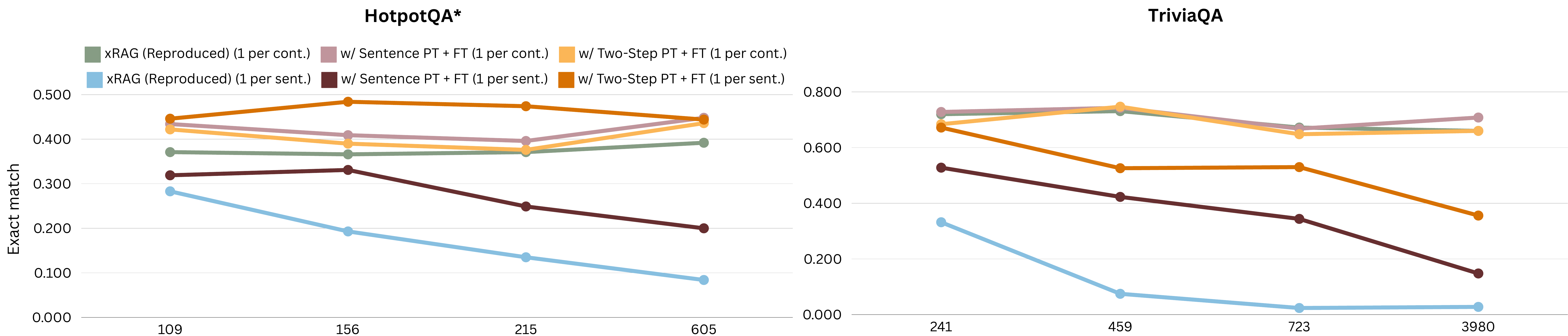}
    \caption{Downstream task performance of different variants of xRAG binned by the length of the input context. Both datasets are evaluated in terms of exact match. Each bucket contains approximately 250 samples.}
    \label{fig:downstream_performance_per_length}
\end{figure*}

\section{Additional Results and Analysis}
\label{app:add_results}

\subsection{Downstream Task Performance}
\label{app:add_results:downstream_tasks}

Tables~\ref{tab:res_xrag_variants_downstream_tasks_extended} and~\ref{tab:res_xrag_variants_downstream_tasks_gsm8k} present additional results for different variants of xRAG method on long-context datasets. Figure~\ref{fig:downstream_performance_per_length} presents downstream tasks performance of different variants of xRAG in terms of length of the context. The following xRAG variants are taken into account:

\paragraph{xRAG (Reproduced)} The pre-training and fine-tuning procedures remain the same as in the original xRAG. We use the same pre-training data. For the fine-tuning step, we use \textasciitilde50\% of the samples they used with original context paragraphs instead of retrieved ones.\footnote{This is due to limitations in the reproducibility of the pre-processing of the fine-tuning data originally used for xRAG.}

\paragraph{xRAG w/ PT + FT (pre-training and fine-tuning with one token per sentence)}  We chunk each sample in pre-training and fine-tuning data into sentences and encode each sentence into a separate xRAG token. 

\paragraph{xRAG w/ Sentence PT + FT (sentence-level pre-training + fine-tuning with one token per sentence)}  pre-training with samples containing single sentences,\footnote{The pre-training step uses only one sentence from each pre-training sample and encodes it as a single xRAG token.} is followed by fine-tuning with samples chunked into sentences with one token per sentence.

\paragraph{xRAG w/ Two-Step PT + FT (sentence-level pre-training + pre-training and fine-tuning with one token per sentence)}  Two-step pre-training involves: (\emph{i})~encoding one sentence per pre-training sample, and (\emph{ii})~chunking samples into sentences with each sentence encoded separately. It is followed by fine-tuning with samples chunked into sentences with one token per sentence.

\begin{table*}[tp]
    \centering    
    \resizebox{0.8\textwidth}{!}{%
    \setlength{\tabcolsep}{5pt}
    \begin{tabular}{lcccccccccccc}
        \toprule
        \multirow{2}{*}{\textbf{Method}} & \multicolumn{2}{c}{\textbf{HotpotQA}} & \multicolumn{2}{c}{\textbf{HotpotQA*}} & \multicolumn{2}{c}{\textbf{arXiv-sum.}} & \multicolumn{2}{c}{\textbf{QuAC}} & \multicolumn{2}{c}{\textbf{TriviaQA}} & \multicolumn{2}{c}{\textbf{GSM8K}} \\\
         & \textbf{1st} & \textbf{Avg} & \textbf{1st} & \textbf{Avg} & \textbf{1st} & \textbf{Avg} & \textbf{1st} & \textbf{Avg} & \textbf{1st} & \textbf{Avg} & \textbf{1st} & \textbf{Avg} \\
        \midrule
        Mistral-7B & 0.86 & 0.80 & 0.80 &  0.75 & 1.00 & 0.97 & 0.94 & 0.93 & 0.81 & 0.78 & 0.58 & 0.50 \\
        xRAG & 0.61 & 0.52 & 0.65 & 0.57 & 0.36	& 0.39 & 0.50 & 0.45 & 0.77 & 0.73 & 0.50 & 0.42 \\
        PISCO & 0.62 & 0.59 & 0.79 & 0.76 & 0.84 & 0.74 & 0.65 & 0.63 & 0.86 & 0.84 & 0.56 & 0.48 \\
        LLMLingua & 0.55 & 0.45 & 0.81 & 0.75 & 0.58 & 0.62 & 0.56 & 0.49 & 0.78 & 0.72 & 0.52 & 0.44 \\
        \bottomrule
    \end{tabular}
    }
    \caption{Grounding scores with respect to the source document/context for responses generated with different methods. All scores are an average across 5 random sets of 100 samples from the original evaluation set (standard deviation ($\sigma$) between sets is \(<0.03\)). ``1st'': grounding score for the first claim detected in the response. ``Avg'': average grounding score for all the claims in the response.}
    \label{tab:res_baselines_grounding_avg_and_first}
\end{table*}

\begin{table*}[tp]
    \centering
    \resizebox{0.8\textwidth}{!}{%
    \begin{tabular}{llcccccccccc}
        \toprule
        \multirow{2}{*}{\textbf{Method}} & \textbf{xRAG} & \multicolumn{2}{c}{\textbf{HotpotQA}} & \multicolumn{2}{c}{\textbf{HotpotQA*}} & \multicolumn{2}{c}{\textbf{arXiv-sum.}} & \multicolumn{2}{c}{\textbf{QuAC}} & \multicolumn{2}{c}{\textbf{TriviaQA}} \\
         & \textbf{Tokens} & \textbf{1st} & \textbf{Avg} & \textbf{1st} & \textbf{Avg} & \textbf{1st} & \textbf{Avg} & \textbf{1st} & \textbf{Avg} & \textbf{1st} & \textbf{Avg} \\
        \midrule
        xRAG (Reproduced) & \multirow{3}{*}{1 per cont.} & 0.55 & 0.50 & 0.66 & 0.58 & 0.40 & 0.38 & 0.51 & 0.43 & \textbf{0.81} & 0.72 \\
         w/ Sentence PT + FT & & \textbf{0.57} & \textbf{0.52} & \textbf{0.69} & \textbf{0.62} & 0.42 & \textbf{0.44} & \textbf{0.56} & \textbf{0.49} & \textbf{0.81} & \textbf{0.73} \\
         w/ Two-Step PT + FT &  & 0.53 & 0.40 & 0.54 & 0.45 & \textbf{0.43} & 0.35 & 0.50 & 0.42 & 0.76 & 0.68 \\
        \midrule
        xRAG (Reproduced) & \multirow{3}{*}{1 per sent} & 0.31 & 0.31 & 0.51 & 0.47 & 0.00 & 0.26 & \textbf{0.48} & \textbf{0.47} & \textbf{0.76} & \textbf{0.71} \\
         w/ Sentence PT + FT & & \textbf{0.48} & \textbf{0.34} & 0.52 & 0.52 & \textbf{0.30} & \textbf{0.46} & \textbf{0.48} & 0.42 & 0.70 & 0.65 \\
         w/ Two-Step PT + FT &  & 0.26 & 0.23 & \textbf{0.54} & \textbf{0.58} & 0.28 & 0.30 & 0.43 & 0.39 & 0.60 & 0.55 \\
        \bottomrule
    \end{tabular}
    }    
    \caption{Grounding scores with respect to the source document/context for responses generated with different methods. All scores are an average across 5 random sets of 100 samples from the original evaluation set ($\sigma<0.02$). ``1st'': grounding score for the first claim detected in the response. ``Avg'': average grounding score for all the claims in the response.}
    \label{tab:res_xrag_variants_grounding_avg_and_first}
\end{table*}

\begin{table*}[tph]
    \centering
    \resizebox{0.8\textwidth}{!}{%
    \begin{tabular}{llccccc}
        \toprule
        \textbf{Method} & \textbf{Encodings} & \textbf{HotpotQA} & \textbf{HotpotQA*} & \textbf{arXiv-sum.} & \textbf{QuAC} & \textbf{TriviaQA} \\
        \midrule
        xRAG & 1 per cont. & 0 & 0 & 10 & 0 & 0 \\
        \midrule
        xRAG (Reproduced) & \multirow{3}{*}{1 per cont.} & 0 & 0 & 65 & 0 & 0 \\
         w/ Sentence PT + FT & & 0 & 0 & 0 & 0 & 0 \\
         w/ Two-Step PT + FT & & 0 & 0 & 5 & 0 & 0 \\
        \midrule
        xRAG (Reproduced) & \multirow{3}{*}{1 per sent} & 56 & 6 & 98 & 27 & 54 \\
         w/ Sentence PT + FT &  & 4 & 0 & 64 & 0 & 7 \\
         w/ Two-Step PT + FT &  & 2 & 0 & 24 & 0 & 0 \\
         \midrule
        PISCO & 8 per cont. & 3 & 8 & 0 & 0 & 11 \\
        \bottomrule
    \end{tabular}
    }    
    \caption{Number of empty responses generated by soft prompting models.}
    \label{tab:xrag_empty_responses_stats}
\end{table*}

\begin{figure*}[th!]
    \centering
    \includegraphics[width=1\linewidth]{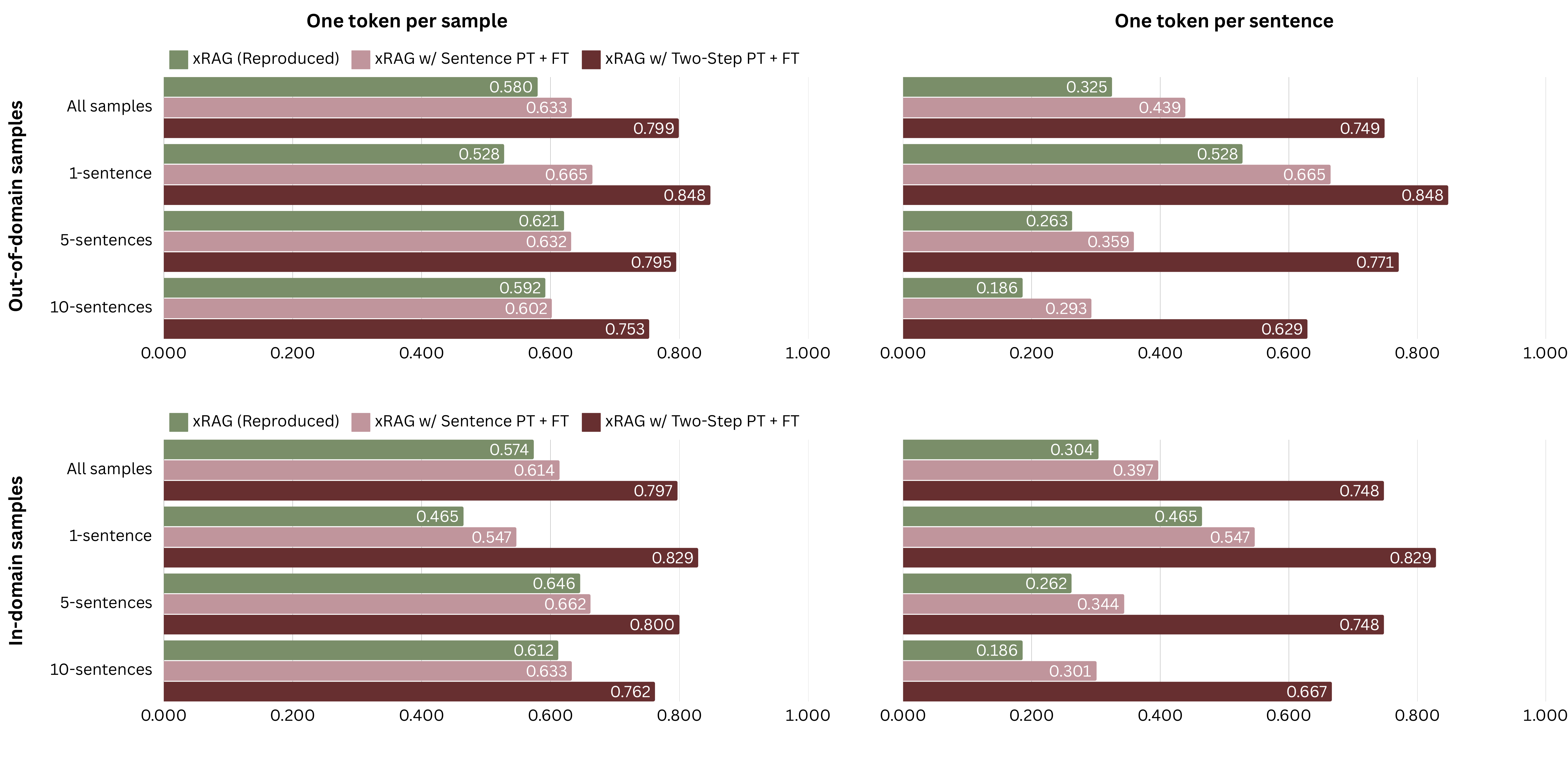}
    \caption{Results of information preservation by different variants of xRAG. Performance is reported in terms of the BERTScore F1 metric computed between the original and reconstructed sample. Results are presented for both in- and out-of-domain samples that are encoded into one token directly or split into sentences and then encoded in multiple tokens.}
    \label{fig:res_xrag_variants_reconstruction_results_two_datasets}
\end{figure*}

\begin{figure*}[th!]
    \centering
    \includegraphics[width=1\linewidth]{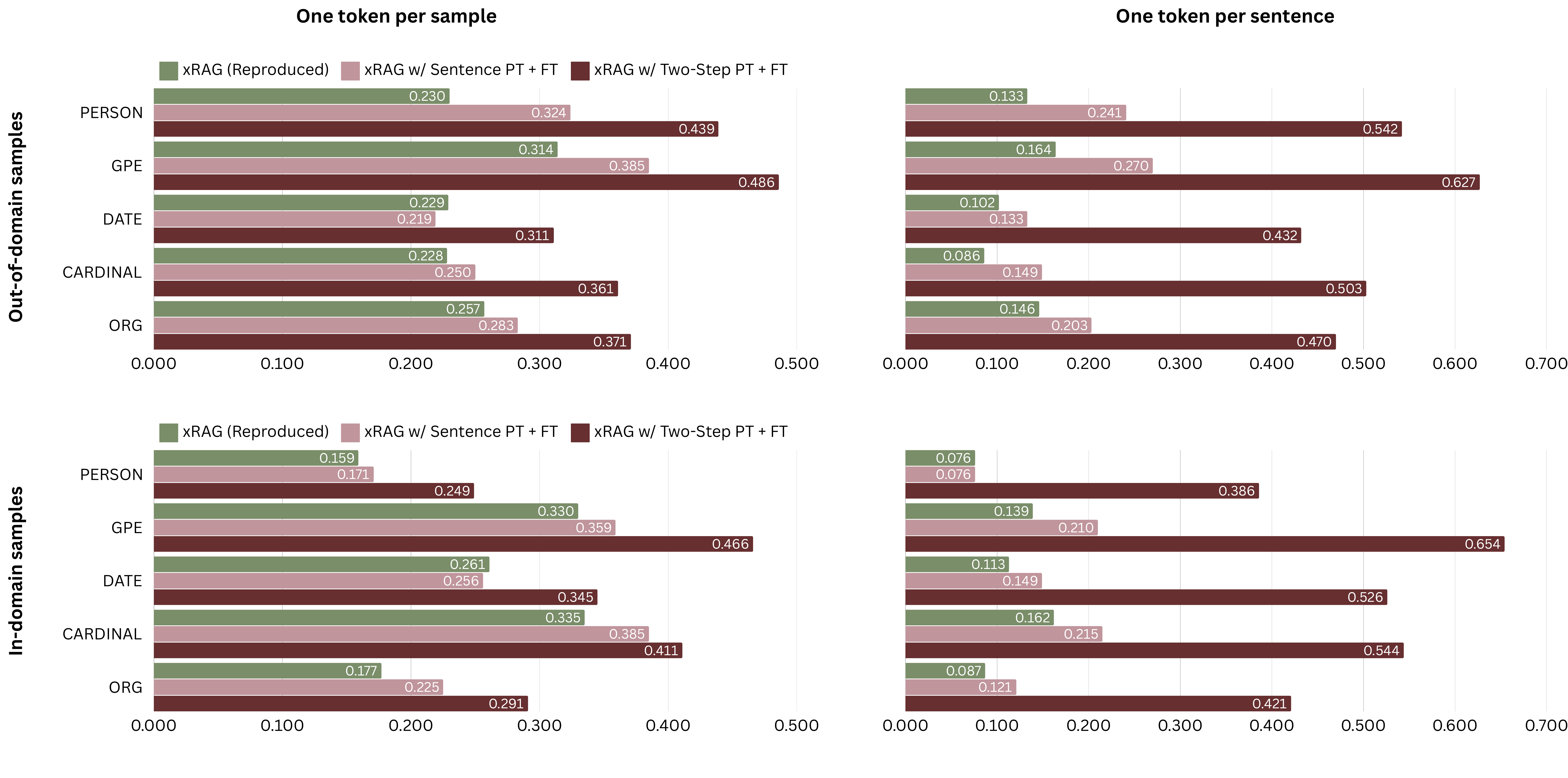}
    \caption{Results of entity preservation experiments. It is evaluated in terms of the exact match between entities and reconstructed text. Results are presented for both in- and out-of-domain samples that are encoded into one token directly or split into sentences and then encoded in multiple tokens.}
    \label{fig:res_entities_preservation_results_two_datasets}
\end{figure*}

\begin{table*}[tph]
    \centering
    \resizebox{0.95\textwidth}{!}{%
    \begin{tabular}{llcccccccc}
        \toprule
        \multirow{2}{*}{\textbf{Method}} & \multirow{2}{*}{\textbf{Encodings}} & \multicolumn{2}{c}{\textbf{Entire dataset}} & \multicolumn{2}{c}{\textbf{1-sent long samples}} & \multicolumn{2}{c}{\textbf{5-sents long samples}} & \multicolumn{2}{c}{\textbf{10-sents long samples}} \\
         &  & \textbf{ROUGE-1} & \textbf{ROUGE-L} & \textbf{ROUGE-1} & \textbf{ROUGE-L} & \textbf{ROUGE-1} & \textbf{ROUGE-L} & \textbf{ROUGE-1} & \textbf{ROUGE-L} \\
        \midrule
        xRAG (Reproduced) & \multirow{3}{*}{1 per cont.} & 0.152 & 0.113 & 0.186 & 0.151 & 0.151 & 0.101 & 0.120 & 0.085 \\
        w/ Sentence PT + FT & & 0.152 & 0.112 & 0.203 & 0.162 & 0.142 & 0.097 & 0.111 & 0.078 \\
        w/ Two-Step PT + FT & & 0.284 & 0.204 & 0.397 & 0.317 & 0.259 & 0.162 & 0.198 & 0.133 \\
        \midrule
        xRAG (Reproduced) & \multirow{3}{*}{1 per sent} & 0.075 & 0.061 & 0.186 & 0.151 & 0.028 & 0.023 & 0.013 & 0.011 \\
        w/ Sentence PT + FT & & 0.091 & 0.072 & 0.202 & 0.162 & 0.043 & 0.032 & 0.027 & 0.021 \\
        w/ Two-Step PT + FT & & 0.384 & 0.273 & 0.397 & 0.317 & 0.434 & 0.294 & 0.320 & 0.209 \\
        \midrule
        \multirow{2}{*}{\tocheck{PISCO}} & 8 per cont. & 0.509 & 0.356 & 0.525 & 0.439 & 0.547 & 0.351 & 0.454 & 0.277 \\
         & 8 per sent. & 0.555 & 0.409 & 0.509 & 0.415 & 0.608 & 0.437 & 0.549 & 0.374 \\
        \bottomrule
    \end{tabular}
    }    
    \caption{Results of information preservation by different variants of xRAG and PISCO on the \textbf{unseen (during training), out-of-domain examples}. Performance is reported in terms of ROUGE metrics computed between the original and reconstructed text. The examples are encoded into one token directly or split into multiple tokens, one per sentence.}
    \label{tab:information_preservation_rouge_unseen}
\end{table*}

\begin{table*}[tph]
    \centering
    \resizebox{0.95\textwidth}{!}{%
    \begin{tabular}{llcccccccc}
        \toprule
        \multirow{2}{*}{\textbf{Method}} & \textbf{Encodings} & \multicolumn{2}{c}{\textbf{Entire dataset}} & \multicolumn{2}{c}{\textbf{1-sent long samples}} & \multicolumn{2}{c}{\textbf{5-sents long samples}} & \multicolumn{2}{c}{\textbf{10-sents long samples}} \\
         &  & \textbf{ROUGE-1} & \textbf{ROUGE-L} & \textbf{ROUGE-1} & \textbf{ROUGE-L} & \textbf{ROUGE-1} & \textbf{ROUGE-L} & \textbf{ROUGE-1} & \textbf{ROUGE-L} \\
        \midrule
        xRAG (Reproduced) & \multirow{3}{*}{1 per cont.} & 0.153 & 0.115 & 0.144 & 0.127 & 0.171 & 0.119 & 0.144 & 0.100 \\
        w/ Sentence PT + FT & & 0.147 & 0.113 & 0.158 & 0.138 & 0.154 & 0.109 & 0.129 & 0.092 \\
        w/ Two-Step PT + FT & & 0.289 & 0.228 & 0.423 & 0.378 & 0.247 & 0.167 & 0.198 & 0.138 \\
        \midrule
        xRAG (Reproduced) & \multirow{3}{*}{1 per sent} & 0.061 & 0.054 & 0.144 & 0.127 & 0.03 & 0.025 & 0.010 & 0.009 \\
        w/ Sentence PT + FT & & 0.080 & 0.066 & 0.158 & 0.138 & 0.048 & 0.035 & 0.034 & 0.026 \\
        w/ Two-Step PT + FT & & 0.422 & 0.327 & 0.423 & 0.378 & 0.451 & 0.315 & 0.391 & 0.287 \\ 
        \midrule
        \multirow{2}{*}{\tocheck{PISCO}} & 8 per cont. & 0.475 & 0.341 & 0.410 & 0.361 & 0.569 & 0.376 & 0.446 & 0.286 \\
         & 8 per sent. & 0.553 & 0.421 & 0.420 &  0.371 & 0.633 & 0.460 & 0.606 & 0.431 \\
        \bottomrule
    \end{tabular}
    }    
    \caption{Results of information preservation by different variants of xRAG and PISCO on the \textbf{seen (during xRAG training), in-domain examples}. The ROUGE metrics are calculated in the same way as for the unseen set in Table~\ref{tab:information_preservation_rouge_unseen}.}
    \label{tab:information_preservation_rouge_seen}
\end{table*}

\subsection{Grounding}
\label{app:add_results:grounding}

Tables~\ref{tab:res_baselines_grounding_avg_and_first} and~\ref{tab:res_xrag_variants_grounding_avg_and_first} report on grounding scores of the base compression methods and our xRAG modifications. In each table we present columns: \emph{1st} -- grounding score of the first claim detected in the response, and \emph{Avg} -- the average grounding score for all the claims in the response. All the grounding scores for xRAG models are reported for the non-empty responses.

Statistics about the number of empty responses returned for each dataset can be found in Table~\ref{tab:xrag_empty_responses_stats}. We can see that the reproduced xRAG model tends to return significantly more empty responses compared to other versions. This is yet another evidence that the newly proposed training regimes lead to more stable representations and better model performance.

\subsection{Information Preservation}
\label{app:add_results:info_preservation}

To better understand the information preservation in the xRAG compression method, we further extend our evaluation with ROUGE metrics (Tables~\ref{tab:information_preservation_rouge_unseen}~and~\ref{tab:information_preservation_rouge_seen}). 
We use the same \emph{seen} (during training, in-domain samples) and \emph{unseen} (during training, out-of-domain samples) datasets as in Section~\ref{subsec:results:infomration_preservation}.
The low ROUGE scores (less than 0.5) indicate the models are not able to reconstruct the original content and instead, they are paraphrasing it, emphasizing the need for a metric that goes beyond lexical matching. Nevertheless, the ROUGE evaluations also show that our xRAG modification (\emph{xRAG w/ Two-Step PT + FT)}) increases more than 2x the ROUGE scores reaching ROUGE-1 values close to 0.4 on the sentence level (both on seen and unseen data). This shows that the new representations are able to capture more precisely the original content compared to the tokens from the reproduced xRAG model.
In addition to entity preservation results for ``\emph{unseen}'' samples discussed in Section~\ref{sec:results}, we provide results for ``\emph{seen}'' samples in Figure~\ref{fig:res_xrag_variants_reconstruction_results_two_datasets}. Both datasets contain 450 examples, out of which 150 samples are one-sentence long, 150 contain 5 sentences and 150 contain 10 sentences. Unseen examples are sampled from three evaluation datasets: HoptpotQA~\citep{Yang:2018:EMNLP}, QuAC~\citep{Choi:2018:EMNLP}, and TriviaQA~\citep{Joshi:2017:ACL} (with 150 examples from each of these datasets).

\paragraph{Named Entity Preservation}

To quantify our observations from the qualitative analysis, we perform additional experiments to measure the amount of entity preservation after compression.
We measure the fraction of text entities from the uncompressed input retained in the reconstructed version (see Figure~\ref{fig:res_entities_preservation_results_two_datasets}). We observe particularly low scores for \emph{dates} and \emph{numerical} values. Additionally, entity preservation for \emph{people} is notably poor on in-domain samples, likely due to noise or ambiguous references in the pre-training data (e.g.,~``\emph{Thomas Scott 1806–1816}'' annotated as PERSON).\footnote{We use \texttt{SpaCy} (\url{https://spacy.io/}) for NER.} In contrast, \emph{geographical locations} are the most consistently preserved across all input types. Overall, our findings indicate that the model struggles with maintaining entities during prompt compression.

In general, the target LLM tends to hallucinate extensively when reconstructing the context, regardless of the amount of encoded information. For example, the following sentence from HotpotQA context: ``\emph{In May 1983, she married Nikos Karvelas, a composer, with whom she collaborated in 1975 and in November she gave birth to her daughter Sofia.}'' is reconstructed as ``\emph{In May 1985, she married Nikos Karvelas, a composer and lyricist.}'' introducing two factual errors (incorrect date and hallucinated information about the lyricist). We must note that different prompts produce widely varying content. Hereby, we do not recommend performing this analysis on a single prompt.

\section{Prompts Used in the Experiments}
\label{app:prompts}

Table~\ref{tab:reconstruction_prompts} contains prompts used in the context reconstruction experiments, Table~\ref{tab:grounding_prompts} contains prompts used for grounding evaluation, and Table~\ref{tab:response_generation_prompts} contains prompts used for response generation for different tasks.

\begin{table}[tp]
    \centering
    \resizebox{\columnwidth}{!}{%
    \setlength{\tabcolsep}{1pt}
    \scriptsize
    \begin{tabular}{lp{6.5cm}}
        \toprule
        \textbf{ID} & \textbf{Prompt} \\
        \midrule
        1 & These two expressions are equivalent in essence:(1)~\emph{\{token\}} (2)\\\midrule
        2 & In other words, background: \emph{\{token\}} is just another way of saying: \\\midrule
        3 & Background: \emph{\{token\}} means the same as \\\midrule
        4 & \emph{\{token\}} After unpacking the ideas in the background information above, we got: \\\midrule
        5 & \emph{\{token\}} Please offer a restatement of the background sentences I've just read. \\
        \bottomrule
    \end{tabular}
    }
    \caption{Prompts used for reconstructing contexts encoded by soft prompt compression method.}
    \label{tab:reconstruction_prompts}
\end{table}

\begin{table*}[tp]
    \centering
    \scriptsize
    \begin{tabular}{lp{12cm}}
        \toprule
         & \textbf{Prompt} \\
        \midrule
        Claim detection & You are trying to verify the faithfulness of statements made in a given summary of an article against the actual text of the article. To do so, you first need to break the summary into a set of "atomic claims", each of which will then be passed to a human who will read the article and verify if the claim is true or not. Each atomic claim must be fully understandable without any other context from the summary (e.g., all entities must be referred to by name, not pronoun), and they must be situated within relevant temporal, location, and causal context whenever possible. Try to keep each atomic claim to a maximum of 2 sentences. Each atomic claim is separated with ’- ’. Summary: {} List of atomic claims: \\
        \midrule
        Faithfulness evaluation & You are provided with a context and a statement. Your task is to carefully read the context and then determine whether the statement is true or false. Use the information given in the context to make your decision. Do not provide explanations. Context: {} Statement: {} Question: Based on the context provided, is the above statement True or False? Answer: \\
        \bottomrule
    \end{tabular}
    \caption{Prompts used for grounding evaluation.}
    \label{tab:grounding_prompts}
\end{table*}

\begin{table*}[tp]
    \centering
    \scriptsize
    \begin{tabular}{lp{12cm}}
        \toprule
        \textbf{Task} & \textbf{Prompt} \\
        \midrule
        QA & [INST] Refer to the background document and answer the question. Provide only a short answer.\textbackslash n\textbackslash nBackground: \{context\}\textbackslash n\textbackslash nQuestion: \{question\} [/INST] The answer is: \\
        \midrule
        Summarization & [INST] Briefly summarize this article:\textbackslash n\textbackslash nArticle: \{context\} [/INST] Summary: \\
        \midrule
        Mathematical reasoning & [INST] Answer the math question by providing a numerical value. Precede the final answer with an explanation of the intermediate steps. Do not add any symbols to the final numerical answer.\textbackslash n\textbackslash n\#\#\#\textbackslash n Here are some examples:\textbackslash n\{icl\_demos\}\textbackslash n\#\#\#\textbackslash n\textbackslash nQuestion: \{question\} [/INST] \\
        \midrule
        Conversational QA & [INST] Refer to the background document, as well as the conversational context and answer the question. Answer the question by extracting a specific span from the provided background.\textbackslash n\textbackslash nBackground: \{context\}\textbackslash n\textbackslash nConversational context: \{conv\_context\}\{question\} [/INST] \\
        \bottomrule
    \end{tabular}
    \caption{Prompts used for response generation for different tasks.}
    \label{tab:response_generation_prompts}
\end{table*}

\begin{table*}[tp]
    \centering
    \scriptsize
    \begin{tabular}{p{1.3cm}p{2.7cm}p{6.2cm}p{3.8cm}}
        \toprule
        \textbf{Method} & \textbf{Prompt} & \textbf{Original text} & \textbf{Reconstructed text} \\
        \midrule
        \multirow{3}{1.3cm}{xRAG} & <xRAG> Please offer a restatement of the background sentences I’ve just read. & In May 1983, she married Nikos Karvelas, a composer, with whom she collaborated in 1975 and in November she gave birth to her daughter Sofia. After their marriage, she started a close collaboration with Karvelas. Since 1975, all her releases have become gold or platinum and have included songs by Karvelas. & In 1985, she married Nikos Karvelas, a composer and music producer. They collaborated on many projects, and she began to write lyrics for him. \\
         & The token <xRAG> represents a compressed text. Please fully expand and reconstruct the original text. & Beginning in 1986, Motel 6 has advertised through radio commercials featuring the voice of writer and National Public Radio commentator Tom Bodett, with the tagline: We\'ll leave the light on for you. & The phrase "We'll leave the light on for you" is a common phrase used by motels to assure potential customers that they will be welcome. The phrase has been used in advertising for the Motel 6 chain since 1962. \\
         & Reconstruct the content of this document: <xRAG>. & Question: A band's members each earn \$20 per gig. If there are 4 members and they've earned \$400, how many gigs have they played? Solution: The band earns \$80 per gig because 4 x 20 = <<4*20=80>>80 They played 5 gigs because 400 / 80 = <<400/80=5>>5 \#\#\#\# 5 & 25 x 40 = 10 \\
        \midrule
        \multirow{3}{1.3cm}{w/ PT + FT} & Background: <xRAG> means the same as & The terrain is mostly flat and the transition from Borovets to the Black Rock takes just over an hour of peaceful walking. & the Black Rock Desert, but is much easier to walk on. \\
         & Background: <xRAG> Can you put the above sentences in your own terms? & The terrain is mostly flat and the transition from Borovets to the Black Rock takes just over an hour of peaceful walking. & The walk from the car park to the Black Rock is a relatively flat and easy one, taking around 1 hour. \\
        & Background: <xRAG> means the same as & Leblois acted unilaterally to send the 22nd Colonial Infantry Regiment to Termes to support the 3rd Colonial Division. & the 22nd Infantry Regiment. \\
        \midrule
        \multirow{3}{1.3cm}{w/ Sentence PT + FT} & These two expressions are equivalent in essence:(\emph{i})~<xRAG> (\emph{ii})~& The letter was thefirst in a series of direct and indirect communications between the White House and the Kremlin throughout the remainder of the crisis. &  The first of a series of letters between the White House and the Kremlin. \\
        & These two expressions are equivalent in essence:(\emph{i})~<xRAG> (\emph{ii})~& The population was 1,871 at the 2010 census. The Flathead Valley, located in Northwestern Montana, is a region of the U.S. state of Montana. & 2010 Census population of 1,871. \\
        & These two expressions are equivalent in essence:(\emph{i})~<xRAG> (\emph{ii})~& The group's only album, Mary Star of the Sea, was released in 2003. &  The Sea Mary is a 2003 album by the group. \\
        \midrule
        \multirow{3}{1.3cm}{w/ Two-Step PT + FT} & <xRAG> After unpacking the ideas in the background information above, we got: & The letter was thefirst in a series of direct and indirect communications between the White House and the Kremlin throughout the remainder of the crisis. &  the first letter in the series of correspondence between the White House and the Kremlin was the first of a series of letters that would be exchanged throughout the remainder of the crisis. \\
        & In other words, background: <xRAG> is just another way of saying: & The population was 1,871 at the 2010 census. The Flathead Valley, located in Northwestern Montana, is a region of the U.S. state of Montana. & 2 The population was 1,870 at the 2010 census. The Flathead Valley is located in the northwest corner of Montana, United States, and is the largest valley in the region. \\
        & In other words, background: <xRAG> is just another way of saying:  & The group's only album, Mary Star of the Sea, was released in 2003. &  The Mary Star, the group's only album, was released in 2003. \\
        \bottomrule
    \end{tabular}
    \caption{Results of reconstructing context using original xRAG model released by the authors of the paper, the xRAG model reproduced by us, and additional new variants of xRAG that we proposed in this paper.}
    \label{tab:res_xrag_reconstruction_examples}
\end{table*}